\documentclass[]{bytedance_seed}

\usepackage[toc,page,header]{appendix}
\usepackage{enumitem}
\usepackage{caption}

\usepackage{minitoc}
\usepackage{algorithm}
\usepackage[noend]{algpseudocode}
\usepackage{amsmath}
\usepackage{amsfonts}
\usepackage{capt-of}
\usepackage{soul}
\usepackage{xcolor}
\usepackage{colortbl}
\usepackage{bm}
\usepackage{pifont}
\usepackage{cancel}
\usepackage{etoc}

\newcommand{\cmark}{\scalebox{1.35}{\textcolor{green!45!black}{\ding{51}}}} 
\newcommand{\xmark}{\scalebox{1.35}{\textcolor{red!60!black}{\ding{55}}}}   
\newcommand{\red}[1]{{\color{red}#1}}

\newcommand{\boldtheta}{\boldsymbol{\theta}}
\newcommand{\boldepsilon}{\boldsymbol{\epsilon}}
\newcommand{\boldc}{\mathbf{c}}
\newcommand{\boldx}{\mathbf{x}}
\newcommand{\boldv}{\mathbf{v}}
\newcommand{\boldg}{\boldsymbol{g}}
\newcommand{\boldI}{\mathbf{I}}
\newcommand{\boldzero}{\boldsymbol{0}}

\definecolor{mygreen}{HTML}{00cf00}
\definecolor{myblue}{HTML}{0000cf}
\definecolor{myorange}{HTML}{e74e00}
\definecolor{mypurple}{HTML}{CC0066}
\definecolor{myred}{HTML}{A60033}

\definecolor{ablation1}{HTML}{EBFEE8}
\definecolor{ablation2}{HTML}{E5F4FB}
\definecolor{ablation3}{HTML}{FFFDEB}
\definecolor{ablation4}{HTML}{EDE5EF}
\definecolor{ablation5}{HTML}{FBE7E7}
\definecolor{ablation6}{HTML}{E6E6FD}
\definecolor{ablation7}{HTML}{FDF7E9}
\definecolor{ablation8}{HTML}{EFE7E4}

\definecolor{cvprblue}{rgb}{0.21,0.49,0.74}
\definecolor{onestepleapblue}{RGB}{79,113,190}
\definecolor{latentconorange}{RGB}{225,131,65}
\definecolor{backproppurple}{RGB}{104,52,155}

\title{ LeapAlign: Post-Training Flow Matching Models at Any Generation Step by Building Two-Step Trajectories  }

\author[1,2,*]{Zhanhao Liang}
\author[2, *, \dagger]{Tao Yang}
\author[2]{Jie Wu}
\author[2]{Chengjian Feng}
\author[1]{Liang Zheng}

\affiliation[1]{The Australian National University}
\affiliation[2]{ByteDance Seed}

\contribution[*]{Equal contribution}
\contribution[\dagger]{Project lead}

\makeatletter
\g@addto@macro\contributionlist{\\[1mm]{\small
zhanhaoliang@outlook.com,
\{yangtao.mry, wujie.10, fengchengjian.cv\}@bytedance.com,
liang.zheng@anu.edu.au
}}
\makeatother

\abstract{
This paper focuses on the alignment of flow matching models with human preferences. 
A promising way is fine-tuning by directly backpropagating reward gradients through the differentiable generation process of flow matching. However, backpropagating through long trajectories results in prohibitive memory costs and gradient explosion. Therefore, direct-gradient methods struggle to update early generation steps, which are crucial for determining the global structure of the final image. 
To address this issue, we introduce LeapAlign, a fine-tuning method that reduces computational cost and enables direct gradient propagation from reward to early generation steps.
Specifically, we shorten the long trajectory into only two steps by designing two consecutive leaps, each skipping multiple ODE sampling steps and predicting future latents in a single step. By randomizing the start and end timesteps of the leaps, LeapAlign leads to efficient and stable model updates at any generation step. 
To better use such shortened trajectories, we assign higher training weights to those that are more consistent with the long generation path. 
To further enhance gradient stability, we reduce the weights of gradient terms with large magnitude, instead of completely removing them as done in previous works. 
When fine-tuning the Flux model, LeapAlign consistently outperforms state-of-the-art GRPO-based and direct-gradient methods across various metrics, achieving superior image quality and image–text alignment. 
}

\date{\today}
\checkdata[Project Page]{\url{https://rockeycoss.github.io/leapalign/}}

\begin{document}
\maketitle

\begin{figure*}[t]
  \centering
  \includegraphics[width=\textwidth]{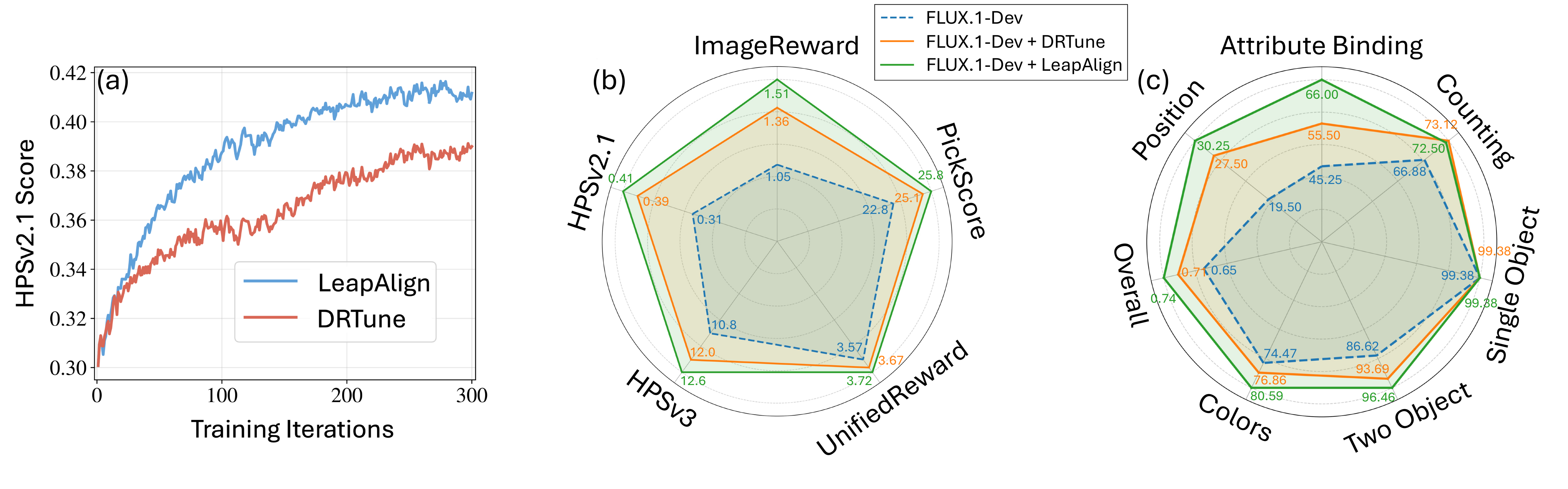}
  \caption{\textbf{Performance overview of LeapAlign.} (a) Comparison of reward improvement during fine-tuning on the compositional alignment task. LeapAlign achieves faster and higher reward gains than DRTune \cite{drtune}. (b) LeapAlign consistently improves Flux across multiple evaluators. (c)~LeapAlign shows clear gains on the GenEval benchmark. For clearer visualization of performance gains, we shift the radar chart origin to 60\% of the FLUX.1-Dev performance and set the maximum radius to the best performance among the displayed methods.}
  \label{fig:performance_overview}
\end{figure*}

\section{Introduction}
\label{sec:intro}
We study how to align flow matching models \cite{liu2022flow,lipman2022flow,esser2024scaling,flux2024,labs2025flux1kontextflowmatching} with human preferences. 
GRPO-based methods, originally designed for large language model (LLM) post-training, are popular for flow matching \cite{dancegrpo,flowgrpo,grpo}. Because the text generation process of LLMs is not differentiable, policy gradient forms the basis of these methods, which inevitably adds a considerable level of stochasticity and variance. 

Essentially, what makes flow matching models differ from LLMs is that the sampling process of the former is continuous and differentiable, while the latter has discrete generation. This difference allows reward gradients to flow through the generation trajectory. That is, to increase the reward, the reward gradient can be backpropagated along intermediate image latents and used to update model weights by the chain rule. We refer to methods using this native gradient-based strategy as \emph{direct-gradient methods}, since they directly backpropagate reward gradients through the differentiable generation trajectory~\cite{imagereward,draft,drtune}. They often allow flow matching post-training to converge faster and train more stably than policy-gradient-based methods.

However, backpropagation through long trajectories poses two significant challenges for direct-gradient methods: 1) prohibitive memory cost caused by the long chains of activations and 2) gradient explosion~\cite{draft}. To avoid these challenges, existing methods typically update only one timestep close to the final image in each iteration~\cite{drtune,draft}. As a consequence, early steps that largely determine image layout \cite{spo,p2pediting} are not updated. While it is possible to enable early-step updates by stopping the gradient at the model input \cite{drtune}, this method discards substantial gradient flow and leads to incomplete optimization. Moreover, while reducing the number of sampling steps may alleviate these problems, it would produce noisy or blurry images, making the rewards predicted by the reward model unreliable. 

In this work, we introduce a flow matching post-training method, LeapAlign, which allows reward gradients to backpropagate to early timesteps while retaining useful gradients. LeapAlign performs training on a \emph{leap trajectory}, a \textit{two-step trajectory} constructed from a standard full-run trajectory, and can fine-tune any generation step.
Specifically, at each iteration, we first sample a full trajectory from noise to image, choose two timesteps $k>j$, and build a leap trajectory that moves the first step from latent $x_k$ to $x_j$ and the second step from $x_j$ to the final latent $x_0$. We compute the reward on the actual final image but backpropagate gradients only through the leap trajectory.  
The leap trajectory keeps the memory cost constant and allows us to directly update any generation step, whether early or late, because $(k,j)$ are randomly selected across the full trajectory. Further, to address gradient explosion and stabilize training, we apply gradient discounting: we down-weight the large-magnitude gradient term instead of removing it, thereby preserving the learning signal that DRTune~\cite{drtune} removes. In addition, we add trajectory-similarity weighting to the loss so that leap trajectories closer to the real path receive higher weights. Together, {LeapAlign} makes early-step fine-tuning practical and stable.

We fine-tune Flux \cite{flux2024} with {LeapAlign} and show the performance gains in Fig.~\ref{fig:performance_overview}. Moreover, compared with the state-of-the-art GRPO-based methods \cite{dancegrpo,li2025mixgrpo} 
and direct-gradient methods \cite{imagereward,draft,drtune}, LeapAlign consistently performs better in image generation, reflected by better scores in HPSv2.1~\cite{hpsv2}, HPSv3~\cite{hpsv3}, PickScore~\cite{pickscore}, UnifiedReward~\cite{unifiedreward}, ImageReward~\cite{imagereward}, and image–text alignment on GenEval~\cite{geneval}. In summary, this paper has the following key points. 

\begin{itemize}[leftmargin=*]
    \item[•] We propose LeapAlign, which trains on a \emph{two-step leap trajectory} carved from a full run. This reduces memory and allows for model updates at any step.
    \item[•] We further propose two techniques for improvement. We assign leap trajectories higher weights if they are similar to the real path. We also scale down gradient terms that potentially have a large magnitude instead of completely removing them to preserve their usefulness.
    \item[•] LeapAlign stably fine-tunes Flux and consistently outperforms existing post-training methods in improving image generation quality and image-text alignment.
\end{itemize}

\section{Related Work}
The emergence of diffusion~\cite{ddpm} and flow matching models~\cite{liu2022flow,lipman2022flow} has driven major progress in text-to-image generation~\cite{rombach2022high,podell2023sdxl,esser2024scaling,flux2024,labs2025flux1kontextflowmatching,seedream2025seedream,wu2025qwen}. 
Aligning such models with human preferences has become increasingly important. 
Inspired by RLHF~\cite{ouyang2022training}, recent studies explore diverse post-training strategies for preference alignment.

Many methods are based on policy gradients \cite{fan2024reinforcement,black2023training,zhang2024large,gupta2025simple,liao2025step,lee2024parrot}. They generally fine-tune 
diffusion models using PPO~\cite{ppo} or REINFORCE~\cite{williams1992simple}.  
Another popular line of work is based on direct preference optimization (DPO)~\cite{dpo} for LLM post-training. They include Diffusion-DPO~\cite{diffusiondpo}, D3PO~\cite{d3po}, SPO~\cite{spo}, and others~\cite{karthik2025scalable,tamboli2025balanceddpo,zhang2024seppo,yuan2024self,li2024aligning,hong2024margin,chen2025towards, densereward, lpo}. They fine-tune diffusion models using preference pairs or sets.
For flow matching models, Adjoint Matching~\cite{domingo2024adjoint} formulates reward fine-tuning as stochastic optimal control, whereas DiffusionNFT~\cite{zheng2025diffusionnft} and AWM~\cite{xue2025advantage} propose forward-process RL methods.
DanceGRPO~\cite{dancegrpo} and Flow-GRPO~\cite{flowgrpo} adapt GRPO~\cite{grpo} to flow matching by converting deterministic ODE sampling into an equivalent SDE formulation and applying the GRPO loss across generation steps. 
MixGRPO~\cite{li2025mixgrpo} and other GRPO variants~\cite{wang2025pref,li2025branchgrpo,zhou2025text} further improve efficiency and performance.

Unlike the methods above, direct-gradient methods use the differentiability of diffusion and flow matching samplers to propagate reward gradients directly \cite{imagereward,draft,drtune,prabhudesai2023aligning,zhang2024itercomp,sorokin2025imagerefl, srpo}. 
ReFL~\cite{imagereward} randomly selects a timestep near the end of the generation trajectory and uses a one-step leap prediction to estimate the final image $\hat{x}_0$. The reward is computed on $\hat{x}_0$, and only the selected step is updated to maximize the reward.
DRaFT-LV~\cite{draft} updates only the last sampling step and reduces gradient variance by repeatedly noising the final image using the forward process and aggregating reward gradients across these noisy variants.
DRTune~\cite{drtune} updates early steps by stopping the gradient at the model input, avoiding out-of-memory errors and gradient explosion when propagating through the full trajectory. 

\textbf{Notable differences with our method.}
Compared with ReFL and DRaFT-LV, which fine-tune only a single late step per trajectory, LeapAlign constructs a leap trajectory (Section~\ref{sec:method:pseudo_traj}) to propagate gradients to early generation steps, which are important for improving global layout. 
Moreover, while DRTune supports early-step updates and can fine-tune multiple steps per rollout, it removes the nested gradient (Section~\ref{sec:method:grad_discounting}), which is useful for capturing dependencies across timesteps. Our method retains this term by lowering its weight in the full gradient, which is shown to be effective. Table~\ref{tab:comparison} compares key differences among these methods. We also summarize these algorithms in Appendix~\ref{appen:algo}. 

\begin{table}[h]
\centering
\caption{Comparison of direct-gradient methods. 
\textit{Early Steps:} whether early generation steps can be updated. 
\textit{Nested Gradient:} whether nested gradients (Eq.~\ref{eq:gradient}) that capture interactions across timesteps are preserved. \textit{Leap Trajectory:} whether the method constructs leap trajectories for backpropogation.
\textit{Multi-Step:} whether multiple steps can be updated per trajectory.} 
\label{tab:comparison}
\begin{tabular}{l|cccc}
\toprule
{Method} & {Early Steps} & {Nested Gradient} & {Leap Trajectory} & {Multi-Step} \\
\midrule
ReFL~\cite{imagereward} & \xmark & \xmark & \xmark & \xmark \\
DRaFT-LV~\cite{draft} & \xmark & \xmark & \xmark & \xmark \\
DRTune~\cite{drtune} & \cmark & \xmark & \xmark & \cmark \\
\textbf{LeapAlign (Ours)} & \cmark & \cmark & \cmark & \cmark \\
\bottomrule
\end{tabular}
\end{table}

\section{Preliminaries}
\textbf{Flow matching models} \cite{lipman2022flow,liu2022flow} learn a continuous transformation that maps Gaussian noise to images by estimating a velocity field.
Let $x_1 \sim \mathcal{N}(\mathbf{0}, \mathbf{I})$ be a Gaussian noise sample and $x_0 \sim p_{\text{data}}$ be a real image from the data distribution.
A forward noising process interpolates between them:
\begin{equation}
\label{eq:flow_matching_noise}
x_t = \alpha_t x_0 + \beta_t x_1,
\end{equation}
where $(\alpha_t, \beta_t)$ is a scheduler~\cite{lipman2024flow} controlling the interpolation from data to noise.

A neural network $v_\theta$ is trained to predict the velocity field
$v = \frac{dx_t}{dt}$ by minimizing:
\begin{equation}
\label{eq:flow_matching_loss}
\mathcal{L}_{\text{fm}}
= \mathbb{E}_{\,t,\,x_0 \sim p_{\text{data}},\,x_1 \sim \mathcal{N}(\mathbf{0}, \mathbf{I})}
\left\lVert v_\theta(x_t, t) - v \right\rVert_2^2.
\end{equation}

In rectified flow matching~\cite{liu2022flow}, the scheduler takes the simple linear form
$\alpha_t = 1 - t,\; \beta_t = t$, making $v = x_1 - x_0$.

\textbf{One-step leap prediction.}
As derived in Appendix~\ref{appen:one_step_leap_derivation}, a rectified flow matching model can estimate the latent $x_j$ at any timestep $j$ from another timestep $k$ by:
\begin{equation}
\label{eq:one_step_skip}
\hat{x}_{j \mid k} = x_k - (k - j)\,v_\theta(x_k, k),
\end{equation}
where $k, j \in [0,1]$. $\hat{x}_{j \mid k}$ could be an approximation of ${x}_{j}$. As detailed in Section \ref{sec:method:pseudo_traj},
this property allows us to construct two adjacent one-step leaps, each directly connecting two timesteps along the full sampling trajectory and thus making backpropagation easier.

\section{Proposed Approach}

\begin{figure*}[htbp]
\centering
\includegraphics[width=1.0\textwidth]{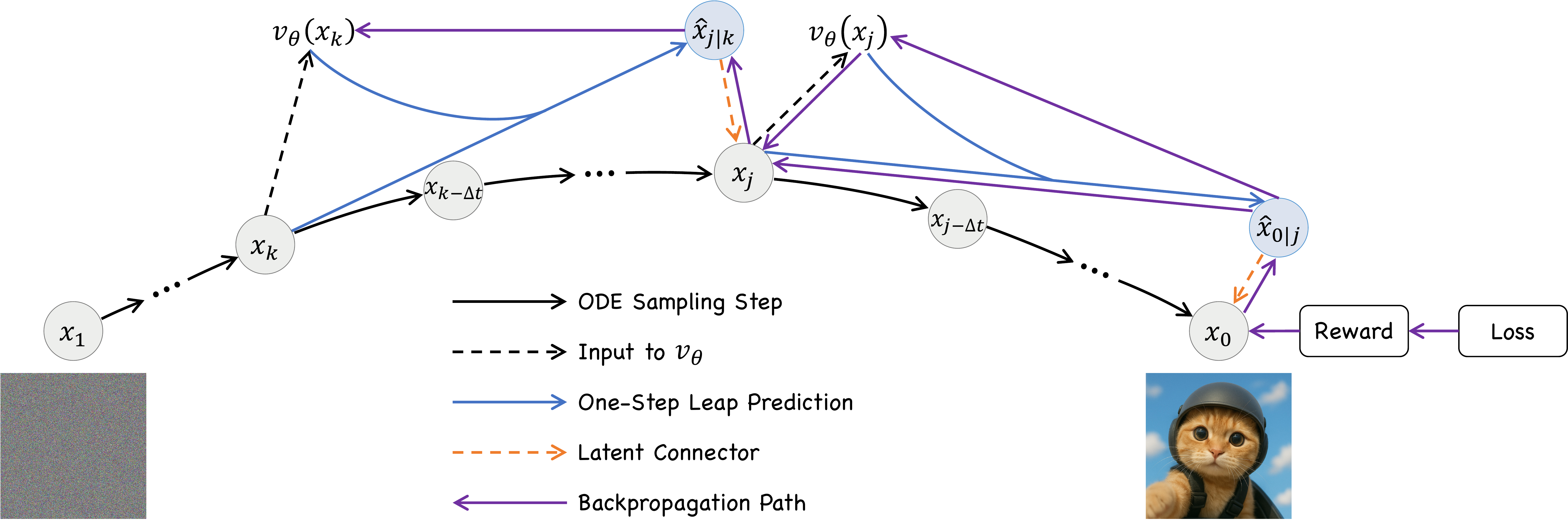}
\caption{Overview of LeapAlign. $x_1,...,x_0$ are the latents in the full generation trajectory, where $x_1$ and $x_0$ correspond to noise and the clean image, respectively. Our method builds two \textcolor{onestepleapblue}{leaps}: from $x_k$ we predict $\hat{x}_{j \mid k}$ using the velocity predicted at $x_k$, and from $x_j$ we predict $\hat{x}_{0 \mid j}$ using the velocity predicted at $x_j$. Here, all latents and velocity predictions are obtained during online sampling. We also compute the \textcolor{latentconorange}{latent connector} to connect the real latent and its one-step approximation. The two \textcolor{onestepleapblue}{leaps} and the two \textcolor{latentconorange}{latent connectors} form a \emph{two-step leap trajectory}\protect\footnotemark. It is along the leap trajectory instead of the full trajectory that reward gradient can \textcolor{backproppurple}{flow} efficiently. Further, because $k$ and $j$ are randomly selected, ultimately LeapAlign can update any generation step.}
\label{fig:method_overview}
\end{figure*}

\footnotetext{The latent connectors, \textit{e.g.}, from $\hat{x}_{j \mid k}$ to $x_j$, are not counted as a step because they do not involve prediction using the flow matching model.}

\subsection{Framework Overview}
To enable effective fine-tuning of early generation steps with direct-gradient methods, we propose {LeapAlign}.
Figure~\ref{fig:method_overview} depicts its overall workflow.
At each iteration, we first generate an image from Gaussian noise through standard ODE sampling steps.
We then randomly select two timesteps ($k$ and $j$) from this long generation trajectory to construct a shortened trajectory of two one-step leaps for fine-tuning (Section~\ref{sec:method:pseudo_traj}).
To prevent gradient explosion during backpropagation through the leap trajectory, {LeapAlign} applies a gradient discounting mechanism (Section~\ref{sec:method:grad_discounting}) that scales down the gradient term with a large norm instead of removing it. 
The fine-tuning objective (Section~\ref{sec:method:objective}) aims to maximize the expected reward of generated images.
Finally, a trajectory-similarity weighting scheme (Section~\ref{sec:method:weighting}) amplifies learning signals from leap trajectories that better match the true generation process. 

\subsection{Leap Trajectory Construction}
\label{sec:method:pseudo_traj}
As shown in Fig. \ref{fig:method_overview}, we shorten the long trajectory into only two steps by constructing two one-step leaps. 
Our design ensures that the shortened trajectory, named \textit{leap trajectory}, preserves the step dynamics of the original trajectory while keeping memory cost constant and controlling gradient growth. Fine-tuning on leap trajectories allows for stable gradient backpropagation to any generation step.

Formally, we randomly select two timesteps $k$ and $j$ from the generation trajectory, where $k > j$. 
Using the one-step leap prediction property of rectified flow models (Eq.~\ref{eq:one_step_skip}), we estimate the latent states at timesteps $j$ and $0$ as:
\begin{equation}
\label{eq:one_step_prediction_j}
\hat{x}_{j \mid k} = x_k - (k - j)v_\theta(x_k),
\end{equation}
\begin{equation}
\label{eq:one_step_prediction_0}
\hat{x}_{0 \mid j} = x_j - jv_\theta(x_j),
\end{equation}
where $x_k$ and $x_j$ denote the latent states at timesteps $k$ and $j$ along the \textit{long} generation trajectory, and $v_\theta$ is the flow matching model being fine-tuned. For simplicity, we let $v_\theta(x_t)$ denote the velocity prediction after classifier-free guidance~\cite{cfgpaper}, and omit the explicit dependence on the text and timestep conditions.

To align the predicted states $\hat{x}$ with the actual ones $x$ while preserving differentiability, we introduce the \textit{latent connector}:
\begin{equation}
\label{eq:straight_through_k}
x_j = \hat{x}_{j \mid k} + \operatorname{stop\_gradient}(x_j - \hat{x}_{j \mid k}), 
\end{equation}
\begin{equation}
\label{eq:straight_through_j}
x_0 = \hat{x}_{0 \mid j} + \operatorname{stop\_gradient}(x_0 - \hat{x}_{0 \mid j}).
\end{equation}
This process constructs a leap trajectory with two steps: 
\[
x_k \;\textcolor{onestepleapblue}{\bm{\rightarrow}}\; (\hat{x}_{j \mid k} \textcolor{latentconorange}{\bm{\dashrightarrow}} x_j) \;\textcolor{onestepleapblue}{\bm{\rightarrow}}\; (\hat{x}_{0 \mid j} \textcolor{latentconorange}{\bm{\dashrightarrow}} x_0),
\]
where solid arrows represent the one-step leap prediction performed by the flow matching model, while dashed arrows denote latent connectors that align the one-step predicted and real latents. 
Because the leap trajectory only has two steps, we achieve efficient gradient backpropagation to early steps with constant memory cost. Moreover, because $k$ and $j$ are randomly selected, we can fine-tune any step.

\subsection{Gradient Discounting}
\label{sec:method:grad_discounting}
While the leap trajectory controls gradient growth, backpropagating through two flow matching steps still produces larger gradients than one-step direct-gradient methods~\cite{imagereward,draft}. 
Appendix~\ref{appen:grad_derivation} shows the gradient propagated from the image $x_0$ \textit{w.r.t}. parameters $\theta$ can be written as:
\begin{equation}
\label{eq:gradient}
\begin{aligned}
\frac{\partial x_0}{\partial \theta} = 
&\underbrace{-\, j\frac{\partial v_\theta(x_j)}{\partial \theta}
-\,(k-j)\frac{\partial v_\theta(x_k)}{\partial \theta}}_{\text{single-step gradients at $k$ and $j$}} \\
&+\, \underbrace{j(k-j)\frac{\partial v_\theta(x_j)}{\partial x_j}
             \frac{\partial v_\theta(x_k)}{\partial \theta}}_{\text{nested gradient}}.
\end{aligned}
\end{equation}

We refer to the first two terms as \emph{single-step gradients}, since each arises from the
gradient of a single one-step leap prediction (Eq.~\ref{eq:one_step_prediction_j} and
Eq.~\ref{eq:one_step_prediction_0}). The last term is the \emph{nested gradient}, which arises when gradients are propagated through multiple steps. The nested gradient is useful for capturing interactions across different generation steps.

DRTune~\cite{drtune} mitigates gradient explosion by stopping the gradient of the model input, which effectively means removing the nested gradient term 
$j(k-j)\frac{\partial v_\theta(x_j)}{\partial x_j}\frac{\partial v_\theta(x_k)}{\partial \theta}$ in Eq.~\ref{eq:gradient}. 
Its drawback is that it loses useful signals in the nested gradient. 
Instead of removing it, we propose a \textit{gradient discounting} mechanism that reduces its magnitude so as to preserve the full gradient structure.

Specifically, using a discounting factor $\alpha \in [0, 1]$, we modify Eq.~\ref{eq:one_step_prediction_0} as:
\begin{equation}
\label{eq:one_step_prediction_0_discounted}
\hat{x}_{0 \mid j} = x_j - jv_\theta(\alpha x_j + (1-\alpha)\operatorname{stop\_gradient}(x_j)).
\end{equation}
This adjustment scales the nested gradient by $\alpha$, producing:
\begin{equation}
\label{eq:discounted_grad}
\begin{aligned}
\frac{\partial x_0}{\partial \theta} = 
&-\, j\frac{\partial v_\theta(x_j)}{\partial \theta}
-\,(k-j)\frac{\partial v_\theta(x_k)}{\partial \theta} \\
&+\, \red{\alpha} j(k-j)\frac{\partial v_\theta(x_j)}{\partial x_j}
             \frac{\partial v_\theta(x_k)}{\partial \theta}.
\end{aligned}
\end{equation}
By adjusting $\alpha$, we can moderate the gradient magnitude without discarding any component of the gradient flow. This, together with the leap trajectory design, stabilizes optimization while retaining full learning signals.

\subsection{Fine-Tuning Objective}
\label{sec:method:objective}

The aim of fine-tuning is to maximize the reward predicted for the generated image. However, directly maximizing reward values often leads to reward hacking, where the model exploits the reward function rather than genuinely improving alignment quality.  
This paper therefore uses a simple hinge-style objective following \citet{imagereward}:
\begin{equation}
\label{eq:raw_loss}
\mathcal{L}_{\text{raw}} = \max\!\left(0, \lambda - r(x_0)\right),
\end{equation}
where $r(\cdot)$ is the reward model, and $\lambda$ is a threshold that controls the strength of reward maximization.  
This loss encourages the model to increase rewards beyond the threshold while preventing unstable optimization toward excessively high or misleading reward values.

Unlike existing direct-gradient methods (\textit{e.g.}, ReFL, DRTune) that effectively estimate rewards from one-step leap predictions (\textit{e.g.}, $\hat{x}_{0|j}$), we evaluate the reward using the generated image $x_0$.  
While $\hat{x}_{0|j}$ is only an estimation of the final output and may contain noise and artifacts, $x_0$ directly reflects the output quality of the full generation trajectory.
Using $x_0$ therefore allows the reward model to make more faithful assessments of visual and semantic quality, providing more reliable supervision signals for fine-tuning.

\subsection{Trajectory-Similarity Weighting}
\label{sec:method:weighting}
Since gradients are backpropagated through the leap trajectory to fine-tune the flow matching model, leap trajectories that deviate significantly from the original generation trajectory can yield misleading gradient signals. 
We therefore further introduce a \emph{trajectory-similarity weighting} that emphasizes leap trajectories that are more consistent with the original trajectory.

We measure similarity by the average absolute difference between predicted states $\hat{x}$ and actual states $x$ at the two connection points:
\[
d_j = \operatorname{mean}\!\left(|x_j - \hat{x}_{j \mid k}|\right), 
d_0 = \operatorname{mean}\!\left(|x_0 - \hat{x}_{0 \mid j}|\right).
\]
To avoid overemphasizing near-identical pairs, we clamp each distance with a minimum value $\tau$ and define the weighting factor as:
\begin{equation}
\label{eq:traj_sim_weighting}
w_{\text{sim}} = \frac{1}{\max(d_j, \tau) + \max(d_0, \tau)}.
\end{equation}
The final objective is formulated as:
\begin{equation}
\label{eq:final_loss}
\mathcal{L} = \operatorname{stop\_gradient}(w_{\text{sim}})\,\mathcal{L}_{\text{raw}}.
\end{equation}
This weighting assigns higher importance to leap trajectories that better match the original generation dynamics, enabling more faithful and effective supervision.

\section{Discussions}
\label{sec:discussion}
\textbf{LeapAlign reflects the key designs from DRTune and ReFL.}
All three methods directly backpropagate reward gradients.
ReFL~\cite{imagereward} uses one-step leap prediction to estimate $\hat{x}_0$ from an intermediate latent, enabling gradient updates at that single timestep. Our method similarly employs one-step leap prediction, but extends it by constructing a leap trajectory from $x_k$ to $x_j$ and then to $x_0$. 
DRTune \cite{drtune} and LeapAlign both propagate gradient to early steps and try to address the nested gradients (but in different ways). 

\textbf{What reward models can be used with LeapAlign, any limitations?} 
LeapAlign can accommodate any differentiable reward model. In our experiments (Section~\ref{exp:main_results}), we show that both CLIP-based~\cite{clip} rewards (HPSv2.1~\cite{hpsv2}, PickScore~\cite{pickscore}) and vision-language-model-based rewards (HPSv3~\cite{hpsv3}) lead to effective fine-tuning results. Extending LeapAlign to non-differentiable rewards, perhaps via differentiable value models~\cite{dai2025vard}, is future work.

\textbf{Is LeapAlign applicable to one-step or few-step image generation models?}
It is less important. While reward gradients can be propagated directly in one-step and few-step methods~\cite{meanflow,salimans2022progressive,geng2023one,yin2024one,zhou2024score} because of the very short trajectories, these models fall short of multi-step methods in image quality and alignment.  As such, it is more important to design fine-tuning methods for multi-step models.

\section{Experiments}
\subsection{Experimental Setup}
\textbf{Training prompt datasets.} 
We conduct experiments on two alignment tasks: general preference alignment and compositional alignment. For general preference alignment, following prior works~\cite{dancegrpo,li2025mixgrpo}, we train on a set of 50,000 prompts sampled from the HPDv2 dataset~\cite{hpsv2}. We also use prompts from MJHQ-30k~\cite{mjhq30k} for training. For compositional alignment, we use the 50,000-prompt dataset~\cite{flowgrpo} generated with the official GenEval scripts~\cite{geneval}. This dataset spans six GenEval task categories, with ratio 7:5:3:1:1:0 for Position, Counting, Attribute Binding, Colors, Two Objects, and Single Object, respectively.

\textbf{Test prompt datasets, evaluation protocols, and metrics.} 
For general preference alignment, we follow the evaluation setup in MixGRPO~\cite{li2025mixgrpo} and generate images using the 400-prompt test set of the HPDv2 dataset. To reduce variance in evaluation, we generate four images per prompt, resulting in a total of 1,600 images. We assess the generated images using six automatic evaluators. Specifically, we employ HPSv2.1~\cite{hpsv2}, HPSv3~\cite{hpsv3}, PickScore~\cite{pickscore}, and ImageReward~\cite{imagereward} to evaluate the degree to which generated images align with human preferences. In addition, we use UnifiedReward-Alignment and UnifiedReward-IQ~\cite{unifiedreward} to assess image–text alignment and overall image quality, respectively. We further construct a 500-prompt test split by randomly sampling from MJHQ-30k~\cite{mjhq30k} to evaluate models fine-tuned on the remaining prompts of the same dataset. 

For compositional alignment, we evaluate on the GenEval benchmark~\cite{geneval}, which consists of six compositional generation tasks: single-object generation, two-object generation, counting, colors, spatial position, and attribute binding. Following the official GenEval evaluation protocol, during testing we generate four images per prompt using its 553-prompt test set and employ the provided rule-based evaluators to automatically determine the correctness of each generated image.

\textbf{Implementation details.} 
We fine-tune FLUX.1-dev~\cite{flux2024}, a state-of-the-art open-source rectified flow matching model capable of generating high-quality images. During fine-tuning, by default we use HPSv2.1~\cite{hpsv2} as the reward model and set the loss threshold $\lambda = 0.55$. We optimize all parameters of the Flux DiT using AdamW~\cite{adamw} with a learning rate of $1\mathrm{e}{-5}$, batch size $64$, weight decay $1\mathrm{e}{-4}$, EMA decay rate $0.995$, $\beta_1 = 0.9$, and $\beta_2 = 0.999$. The model is trained for 300 iterations on 16 GPUs. 
For online rollouts during training, we generate images at a resolution of $720 \times 720$ using 25 steps and a classifier-free guidance scale of 3.5~\cite{cfgpaper}. For evaluation, we sample images with the same resolution, 50 steps, and the same guidance scale. 
For our method LeapAlign, we set $\tau = 0.1$ empirically. Since DRaFT-LV~\cite{draft} and DRTune~\cite{drtune} do not have official implementations, we reproduce them based on the pseudo-code provided in their papers. We also adapt the official implementation of ReFL~\cite{imagereward} to Flux for comparison. For additional implementation and training details, please refer to Appendix~\ref{appen:additional_imple_details}.

\subsection{Main Results}
\label{exp:main_results}

\setlength{\tabcolsep}{0.38mm}
\begin{table*}[t]
    \centering
    \footnotesize
    \caption{Comparing different post-training methods. The base model is Flux.
    For the general preference alignment experiments, all post-training methods except MixGRPO use HPSv2.1 as the reward model, so the metric based on HPSv2.1 is marked as `in-domain'. 
    $^\dagger$~Fine-tuned using HPSv2.1, PickScore, and ImageReward as reward models for general preference alignment experiments. $^*$~Implemented by us due to the absence of an official implementation. $^\ddagger$~Adapted to Flux by us from the official implementation. Best scores are in \textbf{bold}, and second-best scores are \underline{underlined}. PS: PickScore; UR: UnifiedReward; IR: ImageReward;  Obj.: Object; Pos: Position; AttrB: Attribute Binding.}
    \begin{tabular}{l|c|ccccc|c|cccccc}
    \toprule
                   & In-Domain & \multicolumn{5}{c|}{Out-of-Domain} & \multicolumn{7}{c}{GenEval Benchmark} \\
    \midrule
       Method & HPSv2.1 $\uparrow$ & HPSv3 $\uparrow$ & PS $\uparrow$ & UR-Align $\uparrow$ & UR-IQ $\uparrow$ & IR $\uparrow$ & Overall & Single Obj. & Two Obj. & Count & Color & Pos & AttrB\\
    \midrule
    \multicolumn{14}{c}{\textit{Pretrained Model}} \\
    \midrule
    Flux & 0.3078 & 13.5020 & 22.7902 & 3.4514 & 3.5708 & 1.0455 & 0.6535 & \underline{99.38} &	86.62 &	66.88 &	74.47 &	19.50 &	45.25 \\
    \midrule
    \multicolumn{14}{c}{\textit{Policy-Gradient-Based Methods}} \\
    \midrule
    DanceGRPO  & 0.3451 & 14.8336 & 23.1186 & 3.4660 & 3.6199 & 1.2347 & 0.6775 & \underline{99.38} & 90.15 & 69.38 & 76.33 & 22.25 & 49.00 \\
    MixGRPO$^\dagger$ & 0.3692 & 14.7530 & 23.5184 & 3.4393 & 3.6241 & \textbf{1.6155} & \underline{0.7232} & \textbf{99.69} & \underline{93.69} & \textbf{80.00} & \underline{80.05} & 24.25 & 56.25 \\
    \midrule
    \multicolumn{14}{c}{\textit{Direct-Gradient Methods}} \\
    \midrule
    ReFL$^\ddagger$ & 0.3852 & 15.5127 & 23.6299 & 3.4786 & 3.6870 & 1.3468 & 0.7011	& \underline{99.38} &	92.68 &	69.06 &	75.80 &	26.75 &	\underline{57.00} \\
    DRaFT-LV$^*$ & 0.3859 & 15.3699 & \underline{23.6437} & \underline{3.4868} & \underline{3.6887} & 1.3384 & 0.7024 &	\textbf{99.69} &	92.42 &	\underline{74.06} &	75.53 &	24.00 &	55.75 \\ 
    DRTune$^*$ & \underline{0.3882} & \underline{15.5606} & 23.5185 & 3.4793 & 3.6679 & 1.3562 & 0.7101	& \underline{99.38} & \underline{93.69} & 73.12 &	76.86 &	\underline{27.50} &	55.50 \\
    \rowcolor{blue!8} \textbf{LeapAlign} & \textbf{0.4092} & \textbf{15.7678} & \textbf{23.7137} & \textbf{3.4984} & \textbf{3.7244} & \underline{1.5104} & \textbf{0.7420}& \underline{99.38} &	\textbf{96.46} &	72.50 &	\textbf{80.59} &	\textbf{30.25} &	\textbf{66.00} \\ 
    \midrule
    \bottomrule
    \end{tabular}
    \label{tab:overall_comparison}
\end{table*}

\textbf{Comparing \underline{general preference alignment} with state-of-the-art post-training methods.}
We use HPSv2.1~\cite{hpsv2} as the reward model and compare LeapAlign with policy-gradient–based methods including DanceGRPO~\cite{dancegrpo} and MixGRPO~\cite{li2025mixgrpo} and direct-gradient methods including ReFL~\cite{imagereward}, DRaFT-LV~\cite{draft}, and DRTune~\cite{drtune}. For DanceGRPO and MixGRPO, we use their official Flux checkpoints on Hugging Face\protect\footnotemark. 
They are trained on the same prompt set for the same number of iterations as ours. Note that DanceGRPO is trained with HPSv2.1 as the reward, while MixGRPO jointly optimizes HPSv2.1, PickScore, and ImageReward. We summarize the results in Table~\ref{tab:overall_comparison}. 

\footnotetext{DanceGRPO: \url{https://huggingface.co/xzyhku/flux_hpsv2.1_dancegrpo}\\ MixGRPO: \url{https://huggingface.co/tulvgengenr/MixGRPO}}

We have the following observations. First, LeapAlign demonstrates strong overall performance, achieving the highest average scores across both in-domain metric HPSv2.1 and out-of-domain metrics HPSv3, PickScore, UnifiedReward-Alignment, and UnifiedReward-IQ. 
Second, while MixGRPO is jointly fine-tuned with three reward models (HPSv2.1, PickScore, and ImageReward), LeapAlign, trained only with HPSv2.1, yields higher average scores on HPSv2.1 and PickScore and remains competitive on ImageReward.
In summary, compared with the state-of-the-art, LeapAlign produces consistent in-domain and out-of-domain reward gains in human preference alignment, image–text consistency, and overall image quality.

\setlength{\tabcolsep}{6mm}
\begin{table}[t]
\centering
\caption{Comparison of post-training methods using various rewards and prompt sets. We fine-tune Flux with PickScore on HPDv2 and with HPSv3 on MJHQ-30k, respectively.}
\label{tab:more_reward}
\begin{tabular}{l|ccc}
\toprule
Method & HPSv2.1 $\uparrow$ & PickScore $\uparrow$ & HPSv3 $\uparrow$\\
\hline
\multicolumn{4}{c}{\textit{Pretrained Model}} \\
\midrule
Flux & 0.3078 & 22.7902 & 10.7624  \\
\midrule
\multicolumn{4}{c}{\textit{Direct-Gradient Methods}} \\
\midrule
ReFL$^\ddagger$ & 0.3852 & \underline{25.2373} & 11.7642  \\
DRaFT-LV$^*$ &  0.3859 & 24.9596 & 11.2701 \\ 
DRTune$^*$ & \underline{0.3882} & 25.1021 & \underline{12.0023} \\
\textbf{LeapAlign} & \textbf{0.4092} & \textbf{25.7589} & \textbf{12.5855}\\ 
\bottomrule
\end{tabular}
\end{table}

\textbf{Effectiveness of LeapAlign with different reward models, prompt sets, and flow matching models.} To further validate the generality of LeapAlign across different reward models and prompt sets, we additionally fine-tune Flux using PickScore on HPDv2 and HPSv3 on MJHQ-30k, respectively. We then evaluate the fine-tuned models on the HPDv2 test set and on a non-overlapping, randomly sampled test split of MJHQ-30k. As shown in Table~\ref{tab:more_reward}, LeapAlign again achieves the best performance across both settings, confirming its robustness. Additional results on SD3.5-M~\cite{esser2024scaling} are provided in Appendix~\ref{appen:sd35_results}, further supporting the generality of LeapAlign across flow matching models.

\textbf{Comparing \underline{compositional alignment} with state-of-the-art post-training methods.}
To verify that LeapAlign effectively fine-tunes early generation steps which largely determine the image layout \cite{p2pediting}, we conduct evaluation on the GenEval benchmark~\cite{geneval} consisting of diverse compositional generation tasks. 
We use HPSv2.1~\cite{hpsv2} as the reward model and adopt the GenEval training prompts from~\citet{flowgrpo} to fine-tune Flux. For DanceGRPO and MixGRPO, we run experiments using their official codebases and recommended hyperparameter settings, with the same HPSv2.1 reward, GenEval training prompt set, and number of training iterations as ours. Results are reported in Table~\ref{tab:overall_comparison}. The GenEval score improvement during fine-tuning for direct-gradient methods is visualized in Appendix~\ref{appen:vis_geneval_improve}.

We observe that LeapAlign outperforms competitive post-training methods by a clear margin, e.g., overall score 0.7420, compared with 0.7232 for MixGRPO, the best policy-gradient-based baseline, and 0.7101 for DRTune, the strongest direct-gradient baseline. The GenEval performance is particularly strong under the `two objects', `colors', `position', and `attribute binding' categories. In fact, MixGRPO can use policy gradients to update early steps, and DRTune is also capable of fine-tuning early steps but discards critical gradients. These results indicate the benefit of fine-tuning early steps and the effectiveness of LeapAlign.

\textbf{Fine-tuning reward curves.} We plot the average HPSv2.1 reward curves during fine-tuning in Fig.~\ref{fig:performance_overview}. Rewards are computed from generated images $x_0$ obtained from rollout trajectories during fine-tuning. Compared with DRTune, LeapAlign exhibits much stronger reward growth.

\begin{figure*}[t]
\centering
\includegraphics[width=\textwidth]{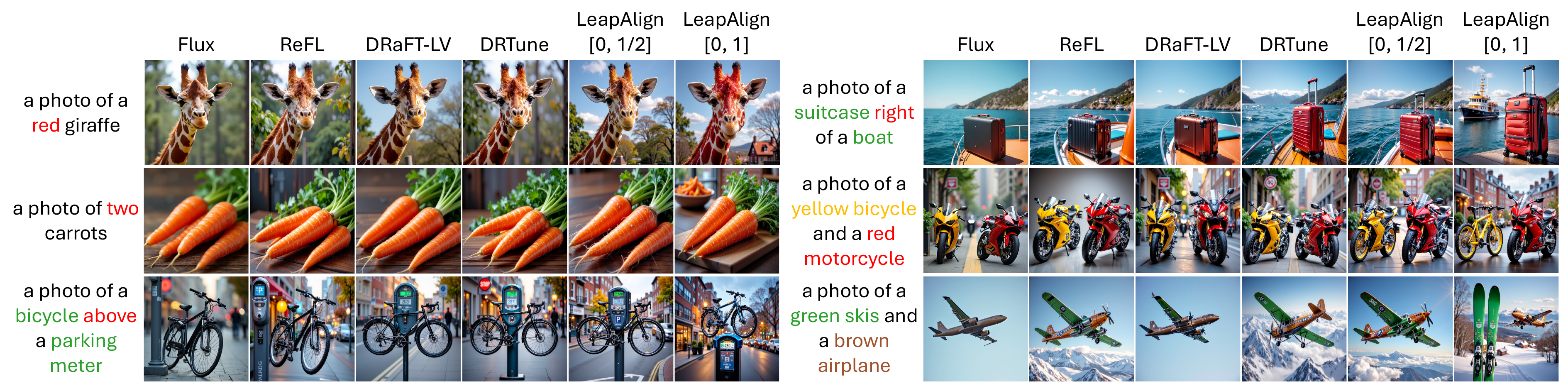}
\caption{Qualitative comparison on GenEval benchmark. We compare direct-gradient post-training methods and the base model Flux. These examples show our method can generate high-quality images aligned with text prompts. $[\,\cdot,\cdot\,]$ indicates the timestep range used for training.}
\label{fig:geneval_qualitative}
\end{figure*}

\textbf{Qualitative results} are shown in Fig.~\ref{fig:geneval_qualitative}. For methods that can only fine-tune late generation steps, such as ReFL and DRaFT-LV, the generated layouts remain similar to those of the pretrained model. In comparison, LeapAlign substantially modifies the global structure, producing images with compositions more faithful to the text prompts.

\subsection{Further Analysis}
\begin{figure*}[t]
  \centering
  \captionsetup{font=footnotesize}
  \captionsetup[subfigure]{font=footnotesize,labelformat=parens,labelsep=space}
  \begin{minipage}[t]{1.0\textwidth}
    \centering
    \begin{subfigure}[t]{0.325\linewidth}
      \centering
      \includegraphics[width=\linewidth]{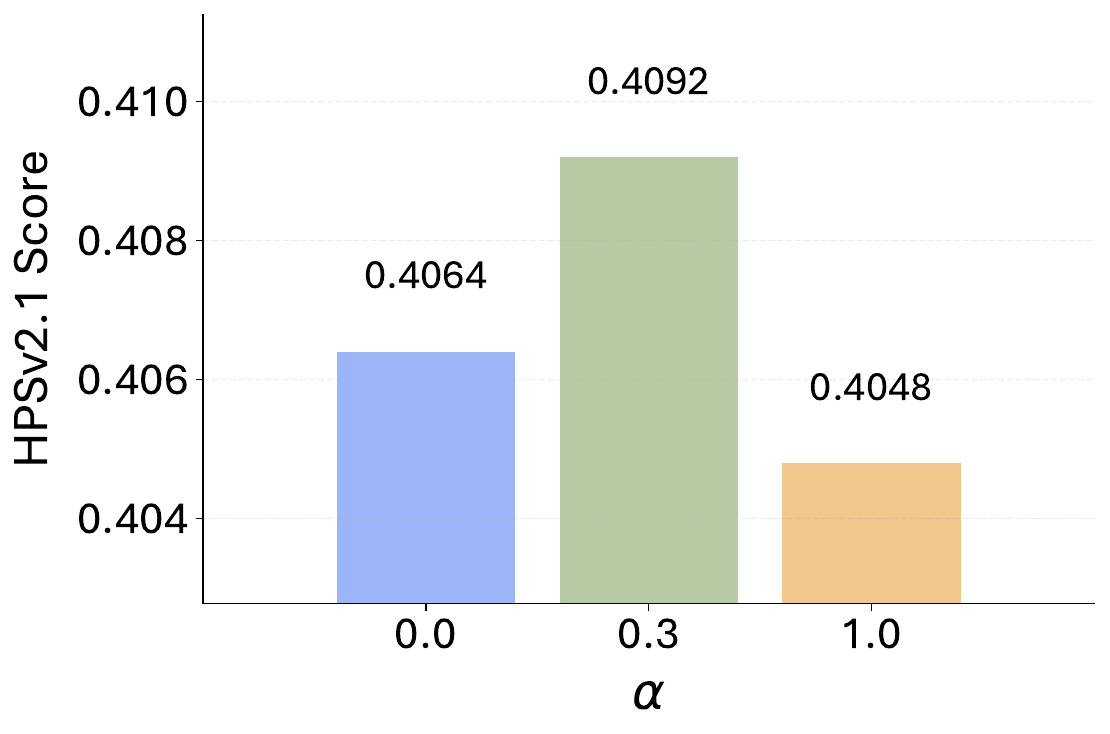}
      \subcaption{\sethlcolor{ablation1}\hl{Effectiveness of gradient discounting.} We vary the value of $\alpha$ (Eq.~\ref{eq:discounted_grad}) and evaluate LeapAlign. Setting $\alpha=0.3$ yields the best performance.}
      \label{fig:grad_discounting}
    \end{subfigure}\hfill
    \begin{subfigure}[t]{0.325\linewidth}
      \centering
      \includegraphics[width=\linewidth]{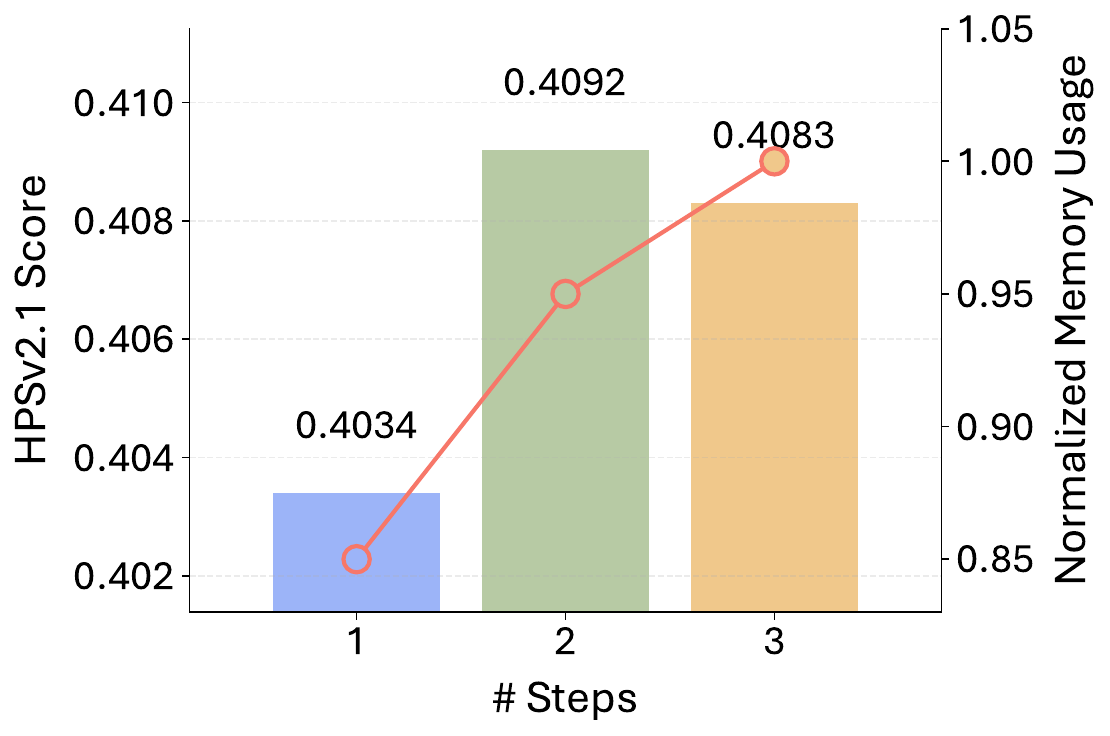}
      \subcaption{\sethlcolor{ablation2}\hl{Comparing using one, two, and three steps in leap trajectories.} Using two steps results in the best trade-off between performance and memory cost.}
      \label{fig:traj_step}
    \end{subfigure}\hfill
    \begin{subfigure}[t]{0.325\linewidth}
      \centering
      \includegraphics[width=\linewidth]{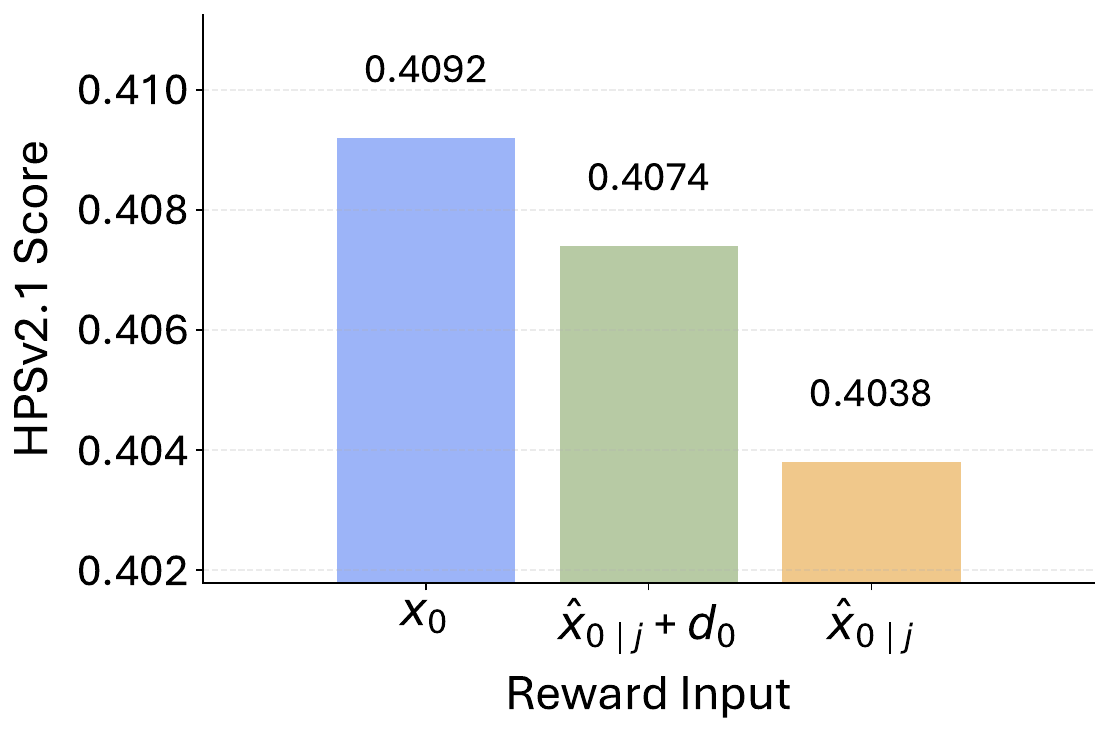}
      \subcaption{\sethlcolor{ablation3}\hl{Comparison of different inputs of the reward model.} `$\hat{x}_{0 \mid j} + d_0$' computes trajectory-similarity weighting with $d_j$ and $d_0$. We can see that using $x_0$ is superior. }
      \label{fig:reward_input}
    \end{subfigure}
    \begin{subfigure}[t]{0.32\linewidth}
      \centering
      \includegraphics[width=\linewidth]{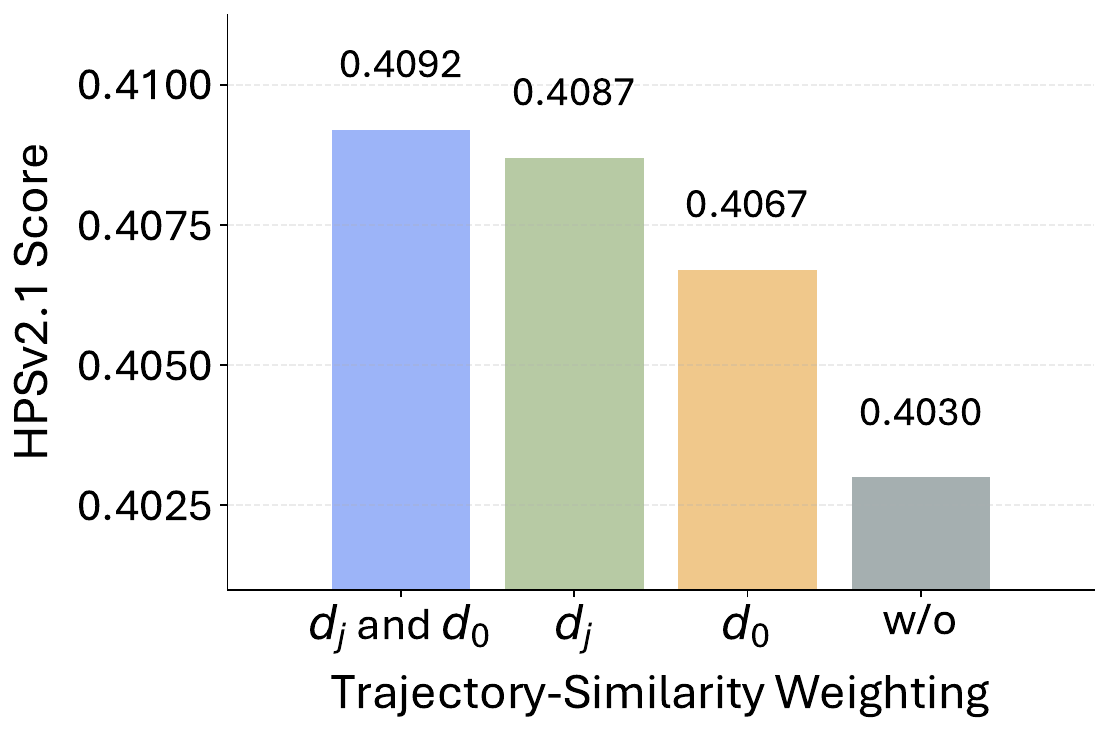}
      \subcaption{\sethlcolor{ablation4}\hl{Comparing trajectory similarity weighting methods.} `$d_j$ and $d_0$': our method (Eq.~\ref{eq:traj_sim_weighting}). `$d_j$' and `$d_0$' only measure trajectory differences at $x_j$ and $x_0$, respectively. `w/o': this mechanism is not applied. Our method has the best result.}
      \label{fig:similarity_weighting}
    \end{subfigure}\hfill
    \begin{subfigure}[t]{0.32\linewidth}
      \centering
      \includegraphics[width=\linewidth]{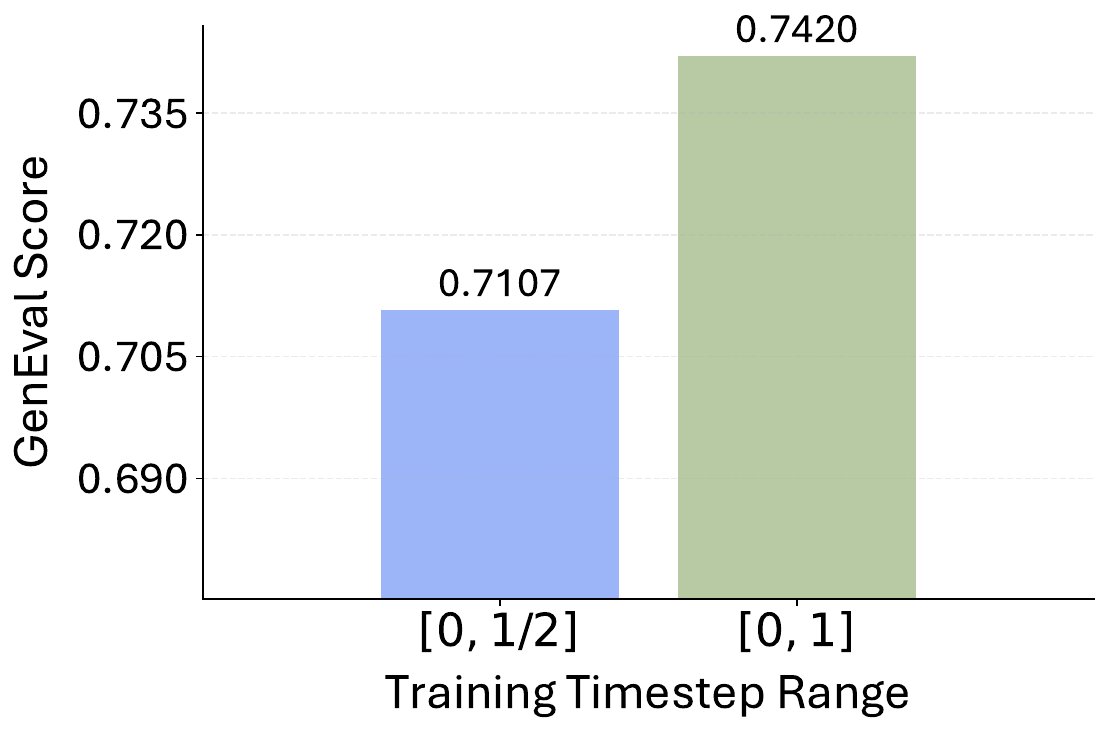}
      \subcaption{\sethlcolor{ablation5}\hl{Impact of training timestep range.} We construct leap trajectories by randomly selecting timesteps from different ranges, where $1$ is the earliest timestep. Random selection over the full timestep range $[0, 1]$ has better performance.}
      \label{fig:t_range}
    \end{subfigure}\hfill
    \begin{subfigure}[t]{0.32\linewidth}
      \centering
      \includegraphics[width=\linewidth]{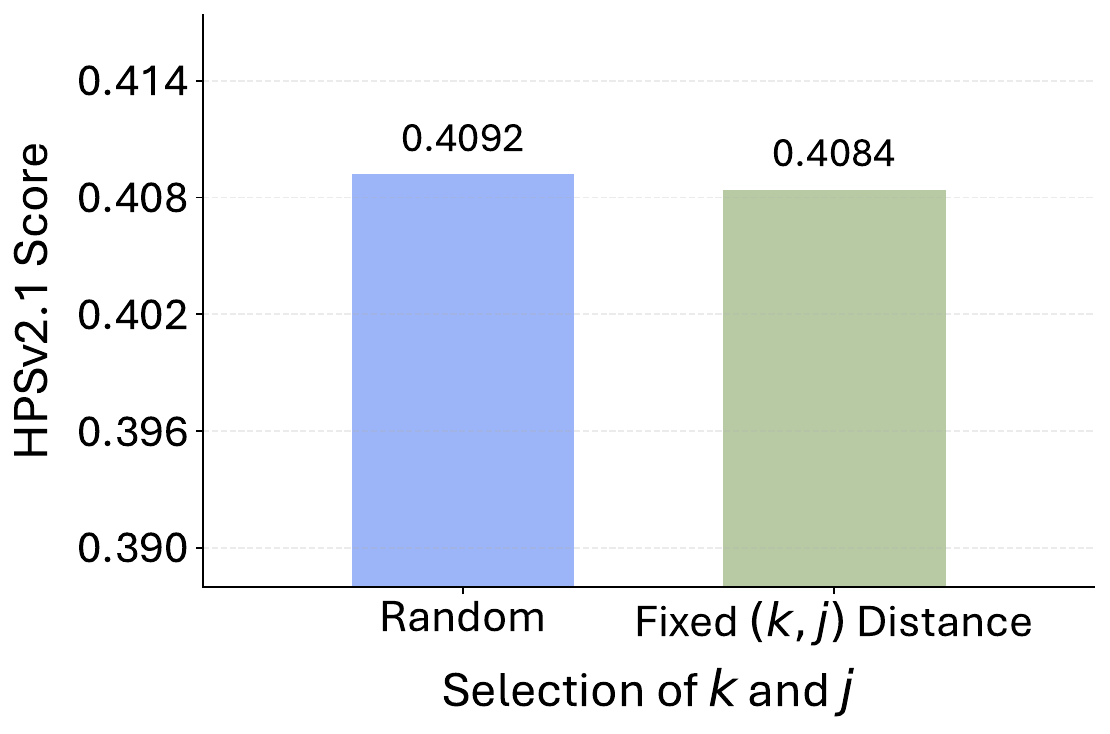}
      \subcaption{\sethlcolor{ablation6}\hl{Comparing strategies for selecting $k$ and $j$.} `Fixed $(k,j)$ Distance': fixing the distance between $k$ and $j$ to $1/2$. `Random' means $k$ and $j$ are randomly selected (Section~\ref{sec:method:pseudo_traj}). Random selection performs better and is easier to implement.}
      \label{fig:t_selection}
    \end{subfigure}
  \end{minipage}
\caption{Further analysis of design components in LeapAlign, including gradient discounting, the number of steps in leap trajectories, the input of the reward model, the trajectory-similarity weighting scheme, the training timestep range, and the selection strategy of $k$ and $j$.}
\end{figure*}

If not specified, we use HPSv2.1 as the reward model when fine-tuning Flux, and adopt the prompt sets from the training and test splits of the HPDv2 dataset~\cite{hpsv2} for training and evaluation, respectively. 

\sethlcolor{ablation1}\hl{\textbf{Effectiveness of gradient discounting.}}
The gradient discounting factor $\alpha$ controls the scale of the nested gradient term (Eq.~\ref{eq:discounted_grad}). To assess its effect, we compare LeapAlign with two variants: one that removes the nested gradient term entirely ($\alpha=0$) and another that applies no discounting ($\alpha=1$). As shown in Fig.~\ref{fig:grad_discounting}, setting $\alpha=0.3$ yields the best performance. Removing the nested gradient term ($\alpha=0$) leads to incomplete optimization and lower scores, while omitting discounting ($\alpha=1$) retains large gradients, making optimization difficult. See Appendix~\ref{appen:additional_analysis} for \emph{additional analysis of the nested gradient}. Notably, even without the nested gradient ($\alpha=0$), LeapAlign still outperforms DRTune on HPSv2.1 (0.4064 vs. 0.3882 in Table~\ref{tab:overall_comparison}), suggesting that its gains come not only from the nested gradient but also from the leap trajectory design.

\sethlcolor{ablation4}\hl{\textbf{Effectiveness of trajectory-similarity weighting.}}
To evaluate the effectiveness of trajectory-similarity weighting, we compare our method (Eq.~\ref{eq:traj_sim_weighting}) with three variants. The first and second variants measure similarity only at $x_j$ and $x_0$, respectively, while the last variant removes the weighting mechanism completely. As shown in Fig.~\ref{fig:similarity_weighting}, variants that consider similarity at only a single step already improve the average HPSv2.1 score over the baseline without weighting. Our design, which incorporates similarity at both $x_j$ and $x_0$, further enhances the average score.

\sethlcolor{ablation2}\hl{\textbf{Comparing leap trajectories with one, two, or three steps.}} We build one, two, or three one-step leaps and compare their fine-tuning performance. Results are shown in Fig.~\ref{fig:traj_step}. Our observation is that two-step leap trajectories provide the best trade-off between performance and memory usage. Using three steps increases memory consumption but yields no better result than our two step version. While using one step is not as good as two steps, its generation quality is still better than competing methods like DRTune and ReFL (Table \ref{tab:overall_comparison}). This demonstrates that the design of LeapAlign, including the leap trajectory (Section~\ref{sec:method:pseudo_traj}), reward evaluation on $x_0$ (Section~\ref{sec:method:objective}), and trajectory-similarity weighting (Section~\ref{sec:method:weighting}), boosts the performance of even the one-step variant.

\sethlcolor{ablation5}\hl{\textbf{Comparing range of training timesteps.}} Timesteps $k$ and $j$ define the position of the two leaps in the full trajectory for training (Section~\ref{sec:method:pseudo_traj}). Our default method is to randomly select $k$ and $j$ within the full timestep range $[0, 1]$, where $1$ is the earliest timestep with Gaussian noise as input. In Fig.~\ref{fig:t_range}, we compare this range with $[0, 1/2]$ on the GenEval benchmark. We find that $[0, 1]$ is superior, highlighting that fine-tuning early generation steps is important for accurate layout and composition. As shown in Fig.~\ref{fig:geneval_qualitative}, the $[0, 1]$ variant also yields qualitatively better results with stronger image-text alignment.

\sethlcolor{ablation6}\hl{\textbf{Comparing strategies for selecting $k$ and $j$.}} This paper uses random selection within range $[0, 1]$. We implement a variant: the distance between $k$ and $j$ is fixed to $\frac{1}{2}$. The two methods are compared in Fig.~\ref{fig:t_selection}, where we observe that random selection is slightly better. For implementation simplicity, we use random selection in LeapAlign. 

\sethlcolor{ablation3}\hl{\textbf{Comparing inputs of the reward model.}} We use the generated image $x_0$ as the input to the reward model. To examine the impact of this choice, we implement two variants: one that uses $\hat{x}_{0 \mid j}$ as input and measures trajectory-similarity only at $x_j$, and another that also uses $\hat{x}_{0 \mid j}$ but applies trajectory-similarity weighting considering similarities at both $x_j$ and $x_0$. As shown in Fig.~\ref{fig:reward_input}, using $x_0$ as input yields superior results, benefiting from more accurate reward evaluation and trajectory-similarity weighting. 

\section{Conclusion}
This paper introduces LeapAlign, a new post-training method that constructs two-step leap trajectories for efficient and stable reward gradient backpropagation. We find it useful to down-scale the large-magnitude gradient term and up-weight leap trajectories that are more similar to the original trajectories. Our method successfully addresses the challenge of propagating reward gradients to early generation steps without incurring excessive memory cost or sacrificing useful gradient terms. This is reflected by consistent improvements over existing post-training methods across a wide range of metrics, including general image preference and image-text alignment. In the future, we will implement and improve LeapAlign in video generation. 

\section{Acknowledgement}
We sincerely thank Jie Liu, Zeyue Xue, Kaiwen Zheng, and Xingjian Leng for insightful discussions.

\clearpage

\bibliographystyle{plainnat}
\bibliography{main}

\clearpage

\beginappendix

\section{Visualization of GenEval Score Improvement During Fine-Tuning for Direct-Gradient Methods}
\label{appen:vis_geneval_improve}
\begin{figure}[h]
\centering
\includegraphics[width=0.5\linewidth]{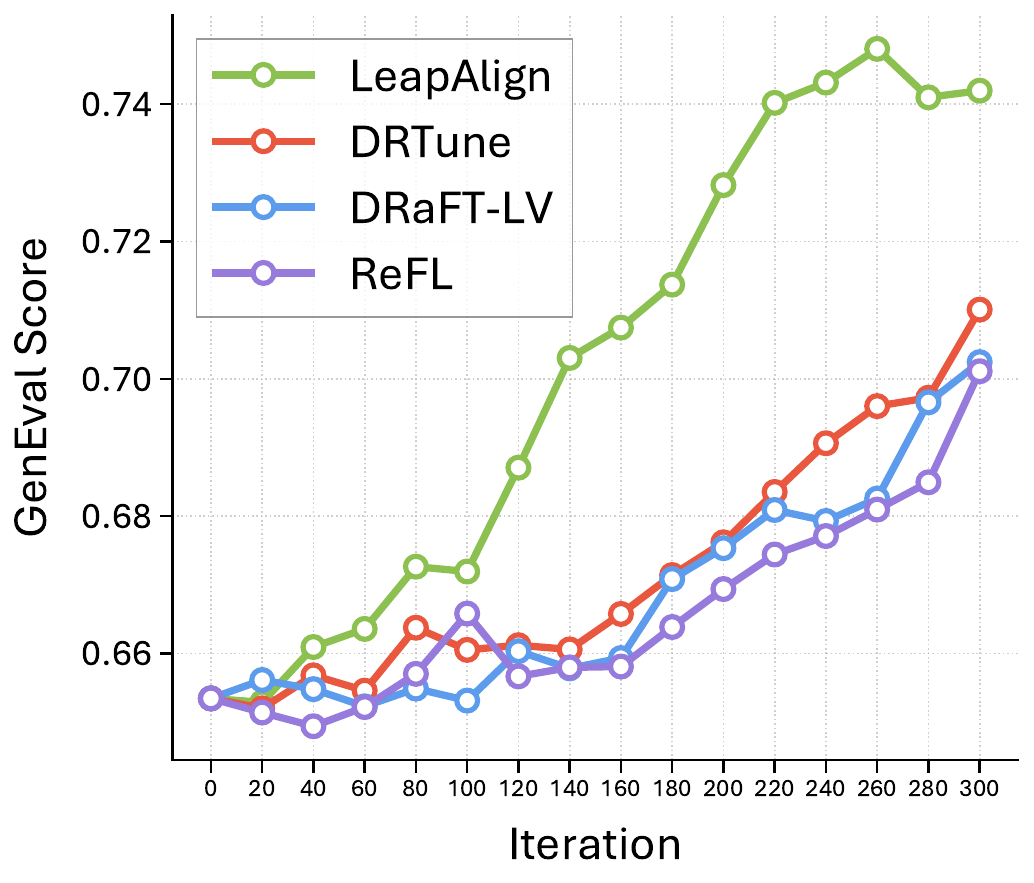}
\caption{Comparison of GenEval score improvement during fine-tuning among ReFL, DRaFT-LV, DRTune, and \textbf{LeapAlign}.}
\label{fig:geneval_improvement}
\end{figure}

Figure~\ref{fig:geneval_improvement} presents the GenEval score improvement curve evaluated during fine-tuning. LeapAlign exhibits both a more rapid increase and a higher final GenEval score compared to DRTune~\cite{drtune}, DRaFT-LV~\cite{draft}, and ReFL~\cite{imagereward}. Methods that update the early generation steps, such as DRTune, achieve stronger improvements than those that do not, underscoring the significance of early-step fine-tuning for the compositional alignment task. LeapAlign optimizes early generation steps more effectively, resulting in the greatest improvement across the entire fine-tuning process.

\section{Additional Results on Stable Diffusion 3.5 Medium}
\label{appen:sd35_results}
To verify that LeapAlign can also achieve strong performance on other flow matching models, we conduct experiments on Stable Diffusion 3.5 Medium~\cite{esser2024scaling}. We fine-tune and evaluate this model at a resolution of $512 \times 512$ for 200 iterations. All other settings follow those used for the general preference alignment task with HPSv2.1 in the main text. Results are shown in Table~\ref{tab:overall_comparison_sd3m}.

We observe that LeapAlign again achieves the best performance across all evaluators compared with other direct-gradient methods. These results demonstrate that LeapAlign generalizes well to other flow matching models and continues to deliver strong improvements.

\setlength{\tabcolsep}{1.0mm}
\begin{table*}[t]
    \centering
    \small
    \caption{Comparison of post-training methods on Stable Diffusion 3.5 Medium. For the general preference alignment experiments, all methods use HPSv2.1 as the reward model, so HPSv2.1 is reported as an `in-domain' metric.
    $^*$~Implemented by us due to the absence of an official implementation. $^\ddagger$~Adapted to Stable Diffusion 3.5 Medium by us from the official implementation. Best scores are in \textbf{bold}, and second-best scores are \underline{underlined}.}
    \begin{tabular}{l|c|ccccc}
    \toprule
                   & In-Domain & \multicolumn{5}{c}{Out-of-Domain} \\
    \midrule
    Method & HPSv2.1 $\uparrow$ & HPSv3 $\uparrow$ & PickScore $\uparrow$ & UnifiedReward-Alignment $\uparrow$ & UnifiedReward-IQ $\uparrow$ & ImageReward $\uparrow$ \\
    \midrule
    \multicolumn{7}{c}{\textit{Pretrained Model}} \\
    \midrule
    SD3.5-M & 0.2967 & 12.2846	& 22.5189 & 3.4436	& 3.5565 & 1.0614 \\
    \midrule
    \multicolumn{7}{c}{\textit{Direct-Gradient Methods}} \\
    \midrule
    ReFL$^\ddagger$ & \underline{0.3833} & 15.2488 & \underline{23.5015} & \underline{3.4810} & \underline{3.6872} & 1.4239 \\
    DRaFT-LV$^*$ &  0.3506 & 14.6022 & 23.0747 & 3.4691 & 3.6519 & 1.3008\\ 
    DRTune$^*$ &  0.3828 & \underline{15.2541} & 23.4711 & 3.4738 & 3.6667 & \underline{1.4320} \\
    \rowcolor{blue!8} \textbf{LeapAlign} & \textbf{0.3915} & \textbf{15.5780} & \textbf{23.6180}	& \textbf{3.4896} & \textbf{3.7182} & \textbf{1.4736} \\ 
    \midrule
    \bottomrule
    \end{tabular}
    \label{tab:overall_comparison_sd3m}
\end{table*}

\section{Additional Analysis}
\label{appen:additional_analysis}
\sethlcolor{ablation8}\hl{\textbf{Analysis of the nested gradient.}}
To better understand the role of the nested gradient, we conduct an additional experiment in which the first step of the leap trajectory is optimized only through the nested gradient. Specifically, we remove the single-step gradient at timestep $k$, as shown in Eq.~\ref{eq:delete_gradient}.
\begin{equation}
\label{eq:delete_gradient}
\begin{aligned}
\frac{\partial x_0}{\partial \theta} = 
&\underbrace{-\, j\frac{\partial v_\theta(x_j)}{\partial \theta}
\cancel{-\,(k-j)\frac{\partial v_\theta(x_k)}{\partial \theta}}}_{\text{single-step gradients at $k$ and $j$}} \\
&+\, \red{\alpha}\underbrace{j(k-j)\frac{\partial v_\theta(x_j)}{\partial x_j}
             \frac{\partial v_\theta(x_k)}{\partial \theta}}_{\text{nested gradient}}.
\end{aligned}
\end{equation}

\begin{figure}[h]
\centering
\includegraphics[width=0.5\linewidth]{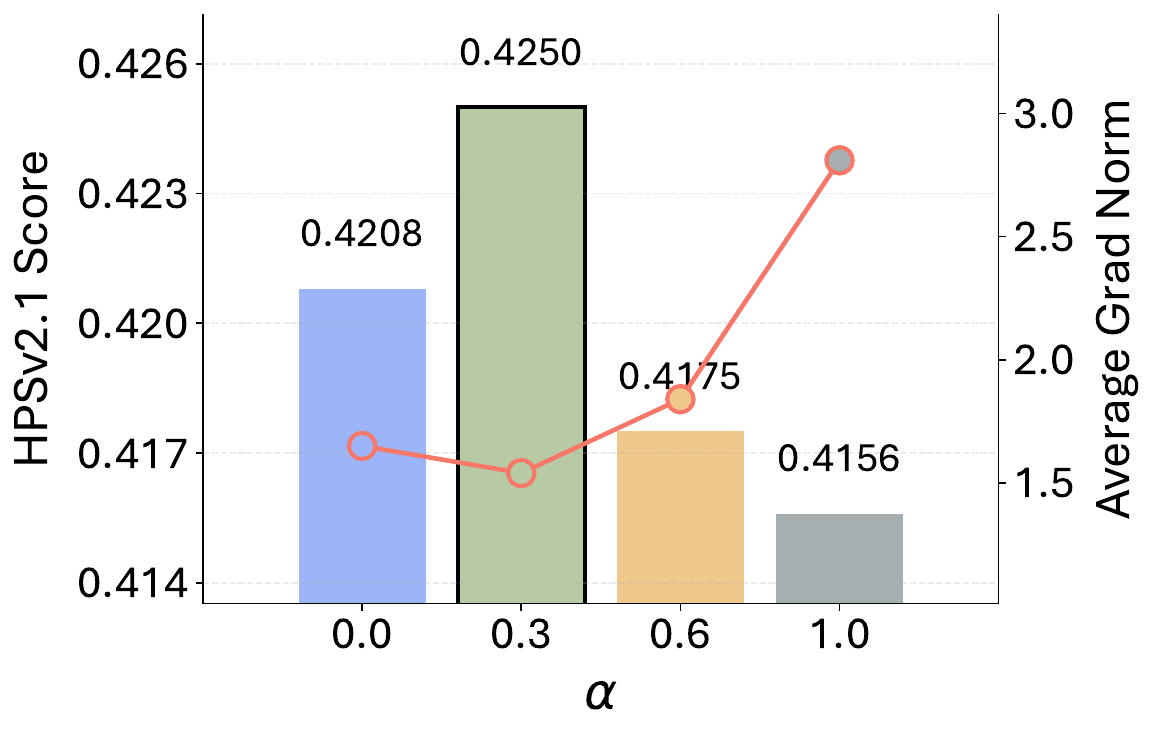}
\caption{\colorbox{ablation8}{\strut \textbf{Analysis of the nested gradient.}} We fine-tune the first step of the leap trajectory using only the nested gradient and vary $\alpha$, which scales its magnitude. Left: average HPSv2.1 score on the test set. Right: average gradient norm during fine-tuning. Directly using the full nested gradient ($\alpha=1$) increases the gradient norm and hurts performance, while moderate gradient discounting factor ($\alpha=0.3$) provides the best trade-off.}
\label{fig:nested_gradient_analysis}
\end{figure}

We fine-tune Flux with HPSv2.1 as the reward model and report results from the non-EMA checkpoint to better expose how the gradient magnitude affects optimization behavior. Fig.~\ref{fig:nested_gradient_analysis} shows the average test-set HPSv2.1 score together with the average gradient norm during fine-tuning. Directly using the full nested gradient ($\alpha=1$) substantially increases the gradient norm and degrades performance compared with removing the nested gradient ($\alpha=0$). In contrast, applying a moderate discount ($\alpha=0.3$) reduces the gradient norm and improves performance over $\alpha=0$. This observation is consistent with the main-paper analysis of gradient discounting and indicates that the nested gradient is beneficial when its magnitude is properly controlled.

\sethlcolor{ablation7}\hl{\textbf{Impact of the loss threshold $\lambda$.}} Table~\ref{tab:lambda_ablation} presents an ablation study on the loss threshold $\lambda$, which controls the strength of reward maximization. When $\lambda$ is too small, the model is under-optimized, resulting in inferior performance. When $\lambda$ is too large, optimization becomes overly aggressive, which hurts out-of-domain generalization and lowers reward scores. Among the tested values, $\lambda=0.55$ achieves the best overall performance, indicating the best trade-off between optimization strength and generalization.

\setlength{\tabcolsep}{6mm}
\begin{table}[h]
\caption{\colorbox{ablation7}{\strut \textbf{Impact of the loss threshold $\lambda$.}}}
\label{tab:lambda_ablation}
\centering
\small
\renewcommand{\arraystretch}{1.1}
\begin{tabular}{l|c|ccc}
\toprule
\ & In-Domain & \multicolumn{3}{c}{Out-of-Domain} \\
\midrule
$\lambda$ & HPSv2.1 & HPSv3 & PickScore & ImageReward \\
\midrule
0.35 & 0.3860 & 15.3635 & 23.4735 & 1.3510 \\
\textbf{0.55} & \textbf{0.4092} & \textbf{15.7678} & \textbf{23.7137} & \textbf{1.5104} \\
0.75 & 0.4091 & 15.7274 & 23.7061 & 1.4844 \\
0.95 & 0.4023 & 15.7254 & 23.5082 & 1.3888 \\
\bottomrule
\end{tabular}
\end{table}

\section{Summary of Direct-Gradient Methods}
\label{appen:algo}
We summarize the direct-gradient methods, including ReFL~\cite{imagereward}, DRaFT-LV~\cite{draft}, DRTune~\cite{drtune}, and our proposed \textbf{LeapAlign}, in Algorithm~\ref{alg:direct-gradient}.

\begin{algorithm}[t]
  \footnotesize
  \caption{Summary of direct-gradient methods}
  \label{alg:direct-gradient}
    \begin{algorithmic}[1]
    \Statex \textbf{Inputs:} pre-trained flow matching model $\boldv_{\boldtheta}$ with parameters $\boldtheta$, reward $r$, prompt dataset $p_{\boldc}$, learning rate $\eta$, early-stop timestep range $m$ ({\color{myorange} ReFL}, {\color{myred} DRTune}), total number of discrete timesteps $T$, number of training
timesteps $K$ ({\color{myred} DRTune}), number of re-noising steps $n$ ({\color{mypurple}DRaFT-LV}), and gradient discounting factor $\alpha$ ({\color{myblue} LeapAlign}).
    \While {not converged}
        \State $t_\text{min} = \begin{cases}
          \text{randint}(1, m)  & \text{\textbf{if} {\color{myorange} ReFL} $\Vert$ {\color{myred} DRTune}} \\
          1  & \text{\textbf{if} {\color{myblue} LeapAlign} $\Vert$ {\color{mypurple} DRaFT-LV} } \\
        \end{cases}$
        \If {{\color{myred} DRTune}}
            \State \smash{$s=\text{randint}(1, T - (K - 1)\lfloor T / K \rfloor)$}
            \State {$t_{\text{train}} = \{s + i \lfloor T / K \rfloor \mid i = 0, 1, \ldots, K - 1\}$}
        \EndIf
        \If{{\color{myblue} LeapAlign}}
            \State {$t_k, t_j \sim \{1, \dots, T\}$ with $t_k > t_j$}
        \EndIf
                \State $\boldc \sim p_{\boldc}$, $\boldx_T \sim \mathcal{N}(\boldzero, \boldI)$   %
        
        \For {$t = T, \dots, 1$}
            \State $\texttt{grad\_on} \gets 
                \begin{cases}
                    t = t_{\min} & \text{\textbf{if} {\color{myorange}ReFL} $\Vert$ {\color{mypurple}DRaFT-LV}}\\
                    t = t_k \Vert t = t_j & \text{\textbf{if} {\color{myblue}LeapAlign}}\\
                    \text{True} & \text{\textbf{if} {\color{myred} DRTune}} \\
                    \text{False} & \text{otherwise}
                \end{cases}
            $
            
            \If{\texttt{grad\_on}}
                \State \texttt{enable\_grad()}
            \Else
                \State \texttt{disable\_grad()}
            \EndIf            
            \If {{\color{myred} DRTune}}
                \State {$v_t = {\boldv_{\boldtheta}}(\texttt{stop\_grad}(\boldx_{t}), t, \boldc)$}
                \If {$t \notin t_{\text{train}}$}
                    \State {$v_t = \texttt{stop\_grad}(v_t)$}
                \EndIf
            \ElsIf {{\color{myblue}LeapAlign} $\texttt{\&\!\&}$ $t = t_j$}
                \State {$\boldx_{t} =\hat{\boldx}_{j \mid k} + \texttt{stop\_grad}(\boldx_{t} - \hat{\boldx}_{j \mid k})$}
                \State {$v_t = \boldv_{\boldtheta}(\alpha\boldx_{t} + (1-\alpha)\texttt{stop\_grad}(\boldx_{t}), t, \boldc)$}
            \Else
                \State {$v_t = \boldv_{\boldtheta}(\boldx_{t}, t, \boldc)$}
            \EndIf
            \If {{({\color{myorange} ReFL} $\Vert$ {\color{myred} DRTune}) $\texttt{\&\!\&}$ $t = t_\text{min}$}}
                \State {$\boldx_0 \approx \texttt{one\_step\_leap\_pred}(\boldx_t, v_t, t, 0)$}
                \State {\textbf{break}}
            \EndIf
            \If {{\color{myblue}LeapAlign}}
                \If {$t = t_k$}
                    \State {$\hat{\boldx}_{j \mid k} = \texttt{one\_step\_leap\_pred}(\boldx_t, v_t, t_k, t_j)$}
                \ElsIf{$t = t_j$}
                    \State {$\hat{\boldx}_{0 \mid j} = \texttt{one\_step\_leap\_pred}(\boldx_t, v_t, t_j, 0)$}
                \EndIf
                \State {$v_t = \texttt{stop\_grad}(v_t)$}
                \State {$\boldx_t = \texttt{stop\_grad}(\boldx_t)$}
            \EndIf
            \State $\boldx_{t-1} = \texttt{step}(\boldx_t, v_t, t)$
        \EndFor
        \State \texttt{enable\_grad()}
        \If {{\color{myblue}LeapAlign}}
            \State \smash{$w=\texttt{stop\_grad}(1/(\texttt{diff}(\boldx_j, \hat{\boldx}_{j \mid k}) + \texttt{diff}(\boldx_0, \hat{\boldx}_{0 \mid j})))$}
            \State {$\boldx_{0} = \hat{\boldx}_{0 \mid j} + \texttt{stop\_grad}(\boldx_{0} - \hat{\boldx}_{0 \mid j})$}
        \Else
            \State {$w=1$}
        \EndIf
        \State $\boldg = -w\nabla_{\boldtheta} r(\boldx_0, \boldc)$ 
        \If {{\color{mypurple} DRaFT-LV}}
            {
            \For {$i = 1, \dots, n$}
                \State $\boldepsilon \sim \mathcal{N}(\boldzero, \boldI)$
                \State $\boldx_1^i = \alpha_1 \texttt{stop\_grad}(\boldx_0) + \beta_1 \boldepsilon$
                \State $\boldx_0^i = \texttt{step}(\boldx_1^i, \boldv_{\boldtheta}(\boldx_1^i, 1, \boldc), 1)$
                \State $\boldg = \boldg - \nabla_{\boldtheta} r(\boldx_0^i, \boldc)$
            \EndFor
            }
        \EndIf
        \State $\boldtheta \gets \boldtheta - \eta \boldg$ 
    \EndWhile
    \State \textbf{return} $\boldtheta$
    \end{algorithmic}
\end{algorithm}

\section{Additional Implementation and Training Details}
\label{appen:additional_imple_details}
During fine-tuning, we set the gradient discounting factor $\alpha$ to 0.3 when using HPSv2.1~\cite{hpsv2}, and to 0.1 when using PickScore~\cite{pickscore} or HPSv3~\cite{hpsv3} as the reward model.
We use a learning rate of $1\mathrm{e}{-5}$ when fine-tuning with HPSv2.1~\cite{hpsv2} or PickScore~\cite{pickscore} as the reward model.
For HPSv3~\cite{hpsv3}, we empirically find that its backpropagated gradients are relatively large, so we adopt a smaller learning rate of $8\mathrm{e}{-6}$.
The loss thresholds $\lambda$ for HPSv2.1, PickScore, and HPSv3 are set to 0.55, 0.4, and 13.5, respectively.

\textbf{Hyperparameters of baseline methods.}
We configure the hyperparameters of baseline direct-gradient methods following the recommended settings in their original papers. Specifically, for DRaFT-LV~\cite{draft}, we set the re-noising steps $n$ to 2. For DRTune, we set the training timesteps $K$ to 2. For both DRTune and ReFL~\cite{imagereward}, we randomly select the early-stop timestep from the last 11 generation steps out of the total 25. We do not include the pre-training loss when fine-tuning with ReFL, as EMA is sufficient to prevent overfitting.

\section{Derivation of the One-Step Leap Prediction}
\label{appen:one_step_leap_derivation}
Let $x_1 \sim \mathcal{N}(\mathbf{0}, \mathbf{I})$ be a Gaussian noise sample and $x_0 \sim p_{\text{data}}$ be a real image drawn from the data distribution. Under a general scheduler $(\alpha_t, \beta_t)$, we can express
\begin{equation}
\label{eq:xtexpression_supp}
x_t = \alpha_t x_0 + \beta_t x_1.
\end{equation}
Following the derivation of \citet{domingo2024adjoint}, the velocity field is defined as
\begin{equation}
\small
\label{eq:vdefinition_supp}
\begin{aligned}
v(x_t, t)
&= \mathbb{E}\!\left[\frac{d x_t}{dt} \,\middle|\, \alpha_t x_0 + \beta_t x_1 = x_t \right] \\
&= \mathbb{E}\!\left[\frac{d\alpha_t}{dt} x_0 + \frac{d\beta_t}{dt} x_1 \,\middle|\, \alpha_t x_0 + \beta_t x_1 = x_t \right].
\end{aligned}
\end{equation}

A simple rearrangement of Eq.~\ref{eq:xtexpression_supp} gives
\[
x_0 = \frac{x_t - \beta_t x_1}{\alpha_t}.
\]
Substituting this into Eq.~\ref{eq:vdefinition_supp} yields
\begin{equation}
\label{eq:vdefinition_no_x0_supp}
\footnotesize
\begin{aligned}
v(x_t, t)
&= \mathbb{E}\!\left[
\frac{d\alpha_t}{dt}\frac{x_t - \beta_t x_1}{\alpha_t}
+ \frac{d\beta_t}{dt} x_1
\,\middle|\, \alpha_t x_0 + \beta_t x_1 = x_t
\right] \\
&= \frac{d\alpha_t}{dt} \frac{x_t - \beta_t \hat{x}_{1\mid t}}{\alpha_t}
+ \frac{d\beta_t}{dt}\, \hat{x}_{1\mid t},
\end{aligned}
\end{equation}
where $\hat{x}_{1\mid t} := \mathbb{E}[x_1 \mid \alpha_t x_0 + \beta_t x_1 = x_t]$. Solving for $\hat{x}_{1\mid t}$ gives
\begin{equation}
\label{eq:predx1_supp}
\hat{x}_{1\mid t} =
\frac{
v(x_t, t) - \frac{d\alpha_t}{dt}\frac{x_t}{\alpha_t}
}{
\frac{d\beta_t}{dt}
- \frac{d\alpha_t}{dt}\frac{\beta_t}{\alpha_t}
}.
\end{equation}

Similarly, rewriting $x_1 = \frac{x_t - \alpha_t x_0}{\beta_t}$ and substituting into Eq.~\ref{eq:vdefinition_supp} gives
\begin{equation}
\label{eq:predx0_supp}
\hat{x}_{0\mid t} =
\frac{
v(x_t, t) - \frac{d\beta_t}{dt}\frac{x_t}{\beta_t}
}{
\frac{d\alpha_t}{dt}
- \frac{d\beta_t}{dt}\frac{\alpha_t}{\beta_t}
}.
\end{equation}

To extend the prediction to an arbitrary timestep $j$, we condition on $x_k$ at timestep $t = k$. Let
\[
\dot{\alpha}_k := \left.\frac{d\alpha_t}{dt}\right|_{t=k},
\qquad
\dot{\beta}_k := \left.\frac{d\beta_t}{dt}\right|_{t=k},
\]
denote the time derivatives of $\alpha_t$ and $\beta_t$ evaluated at $t=k$, and let $v(x_k, k)$ be the velocity at $x_k$. The one-step leap prediction is then
\begin{equation}
\label{eq:predxj_supp}
\footnotesize
\begin{aligned}
\hat{x}_{j\mid k}
&= \alpha_j \hat{x}_{0\mid k} + \beta_j \hat{x}_{1\mid k} \\
&= \alpha_j
   \left[
      \frac{
         v(x_k, k) - \dot{\beta}_k \frac{x_k}{\beta_k}
      }{
         \dot{\alpha}_k - \dot{\beta}_k \frac{\alpha_k}{\beta_k}
      }
   \right]
 + \beta_j
   \left[
      \frac{
         v(x_k, k) - \dot{\alpha}_k \frac{x_k}{\alpha_k}
      }{
         \dot{\beta}_k - \dot{\alpha}_k \frac{\beta_k}{\alpha_k}
      }
   \right].
\end{aligned}
\end{equation}

Under rectified flow matching~\cite{liu2022flow}, the scheduler takes the form
\[
\alpha_t = 1 - t, \qquad \beta_t = t,
\]
so that
\[
\dot{\alpha}_k = -1, \qquad \dot{\beta}_k = 1.
\]
Substituting these into Eq.~\ref{eq:predxj_supp} yields the simplified expression
\begin{equation}
\label{eq:predxj_rflow_supp}
\hat{x}_{j\mid k} = x_k - (k - j)\, v(x_k, k).
\end{equation}

Finally, with a pretrained flow matching model $v_\theta(x_k, k) \approx v(x_k, k)$, the practical one-step leap prediction becomes
\begin{equation}
\label{eq:predxj_rflow_theta_supp}
\hat{x}_{j\mid k} = x_k - (k - j)\, v_\theta(x_k, k).
\end{equation}

\section{Derivation of the Backpropagated Gradient Through the Leap Trajectory}
\label{appen:grad_derivation}
Let $k$ and $j$ be two randomly selected timesteps from the full generation trajectory with $k > j$. The forward pass of the leap trajectory without gradient discounting is
\begin{equation}
\label{eq:leap_traj_forward1}
\hat{x}_{j \mid k} = x_k - (k - j)\, v_\theta(x_k),
\end{equation}
\begin{equation}
\label{eq:leap_traj_forward2}
x_j = \hat{x}_{j \mid k} + \operatorname{stop\_gradient}(x_j - \hat{x}_{j \mid k}),
\end{equation}
\begin{equation}
\label{eq:leap_traj_forward3}
\hat{x}_{0 \mid j} = x_j - j\, v_\theta(x_j),
\end{equation}
\begin{equation}
\label{eq:leap_traj_forward4}
x_0 = \hat{x}_{0 \mid j} + \operatorname{stop\_gradient}(x_0 - \hat{x}_{0 \mid j}).
\end{equation}
In the derivation below, the rollout states $x_k$, $x_j$, and $x_0$ from the full trajectory are treated as detached constants, and gradients are propagated only through the leap trajectory.

The gradient of the final image $x_0$ with respect to the parameters $\theta$ is
\begin{equation}
\label{eq:dx0dtheta}
\small
\begin{aligned}
\frac{\partial x_0}{\partial \theta}
&= \frac{\partial x_0}{\partial \hat{x}_{0 \mid j}} \frac{\partial \hat{x}_{0 \mid j}}{\partial \theta} \\
&= \frac{\partial x_0}{\partial \hat{x}_{0 \mid j}}
\left(
    - j \frac{\partial v_\theta(x_j)}{\partial \theta}
    + \frac{\partial x_j}{\partial \theta}
    - j \frac{\partial v_\theta(x_j)}{\partial x_j} \frac{\partial x_j}{\partial \theta}
\right),
\end{aligned}
\end{equation}
and
\begin{equation}
\label{eq:dxjdtheta}
\begin{aligned}
\frac{\partial x_j}{\partial \theta}
&= \frac{\partial x_j}{\partial \hat{x}_{j \mid k}} \frac{\partial \hat{x}_{j \mid k}}{\partial \theta} \\
&= \frac{\partial x_j}{\partial \hat{x}_{j \mid k}}
\left( - (k-j)\, \frac{\partial v_\theta(x_k)}{\partial \theta} \right).
\end{aligned}
\end{equation}
Since $\frac{\partial x_0}{\partial \hat{x}_{0 \mid j}} = 1$ and $\frac{\partial x_j}{\partial \hat{x}_{j \mid k}} = 1$, substituting Eq.~\ref{eq:dxjdtheta} into Eq.~\ref{eq:dx0dtheta} gives
\begin{equation}
\label{eq:dx0dthetafinal}
\begin{aligned}
\frac{\partial x_0}{\partial \theta}
&= - j \frac{\partial v_\theta(x_j)}{\partial \theta}
   - (k-j)\, \frac{\partial v_\theta(x_k)}{\partial \theta}
   \\ & + j (k-j)\, \frac{\partial v_\theta(x_j)}{\partial x_j}
               \frac{\partial v_\theta(x_k)}{\partial \theta}.
\end{aligned}
\end{equation}

When gradient discounting is applied with gradient discounting factor $\alpha \in [0, 1]$, we modify
Eq.~\ref{eq:leap_traj_forward3} as
\begin{equation}
\label{eq:leap_traj_forward3_discount}
\hat{x}_{0 \mid j}
= x_j - j\, v_\theta\bigl(\alpha x_j + (1-\alpha)\operatorname{stop\_gradient}(x_j)\bigr).
\end{equation}
In the forward pass we still have
$v_\theta(\alpha x_j + (1-\alpha)\operatorname{stop\_gradient}(x_j)) = v_\theta(x_j)$,
but during backpropagation the gradient flowing through
$\frac{\partial v_\theta(x_j)}{\partial x_j}$ is scaled by a factor of $\alpha$,
since
\[
\frac{\partial\bigl(\alpha x_j + (1-\alpha)\operatorname{stop\_gradient}(x_j)\bigr)}{\partial x_j}
= \alpha.
\]
As a result, Eq.~\ref{eq:dx0dthetafinal} becomes
\begin{equation}
\label{eq:dx0dthetafinal_discount}
\begin{aligned}
\frac{\partial x_0}{\partial \theta}
&= - j \frac{\partial v_\theta(x_j)}{\partial \theta}
   - (k-j)\, \frac{\partial v_\theta(x_k)}{\partial \theta} \\
&\quad + \textcolor{red}{\alpha}\, j (k-j)\,
   \frac{\partial v_\theta(x_j)}{\partial x_j}
   \frac{\partial v_\theta(x_k)}{\partial \theta}.
\end{aligned}
\end{equation}

\section{Additional Qualitative Results on the GenEval Benchmark}
\label{appen:geneval_qualitative}
\begin{figure*}[t]
\centering
\includegraphics[width=0.84\textwidth]{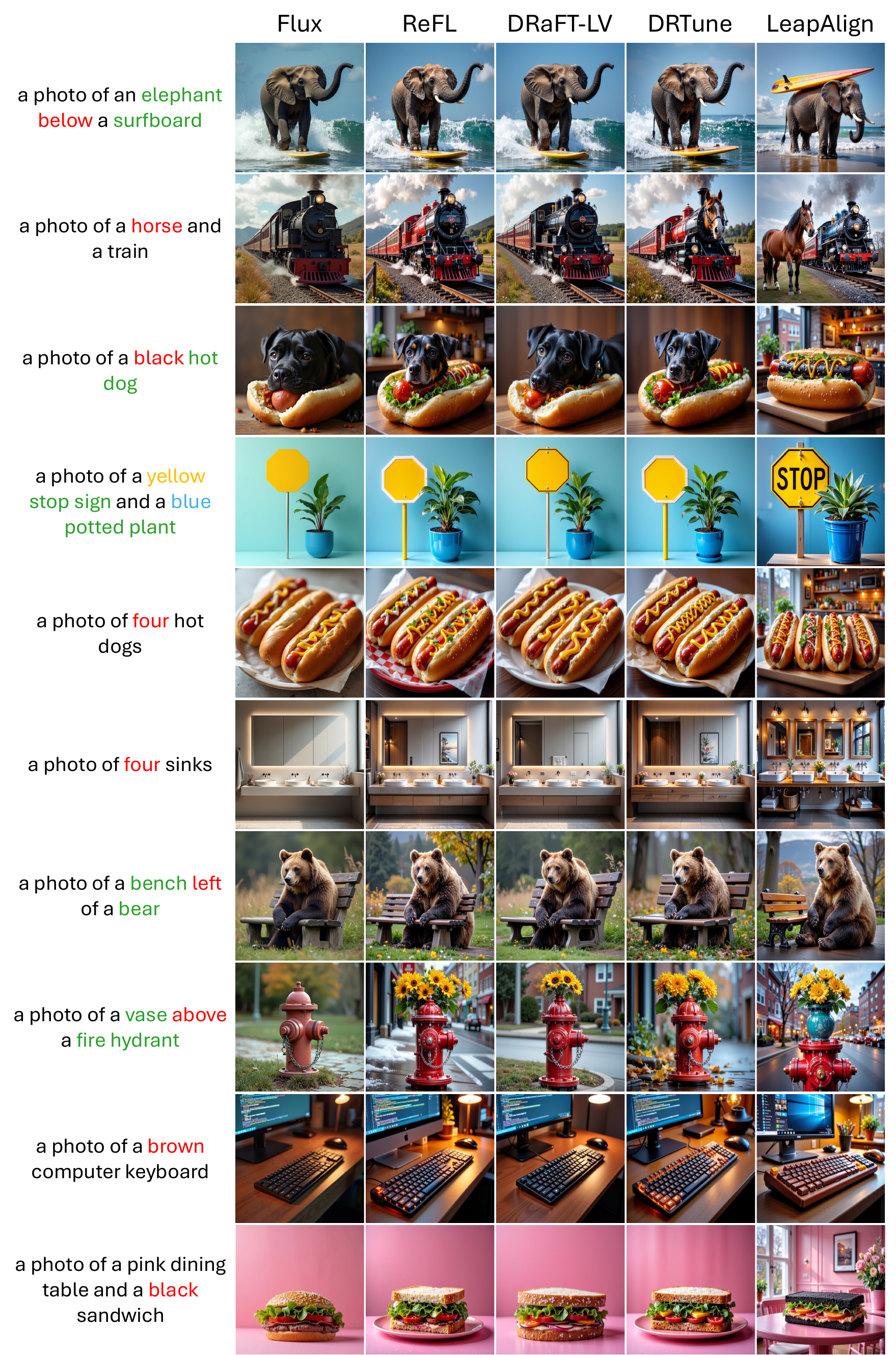}
\caption{Additional qualitative comparisons on the GenEval benchmark.}
\label{fig:more_geneval_qualitative}
\end{figure*}

We present additional qualitative comparisons on the GenEval benchmark across the pretrained Flux model, ReFL, DRaFT-LV, DRTune, and LeapAlign. As shown in Figure~\ref{fig:more_geneval_qualitative}, LeapAlign more effectively adjusts the global structure of the generated images, leading to outputs that more faithfully follow the text prompts.

\section{Qualitative Results of Flux Fine-Tuned with LeapAlign}
\label{appen:qualitative_hpsv3}
\begin{figure*}[t]
\centering
\includegraphics[width=0.84\textwidth]{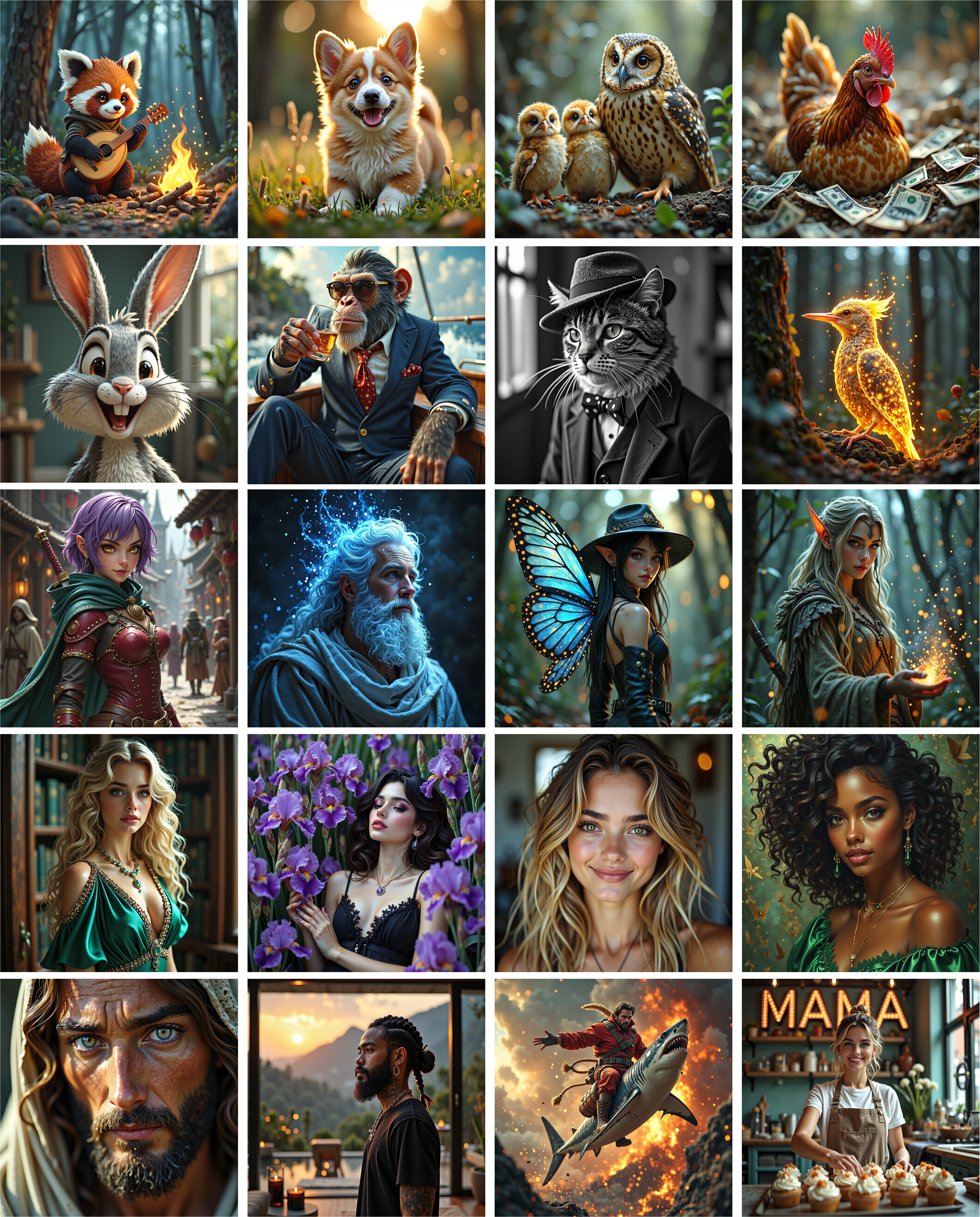}
\caption{Qualitative results of Flux fine-tuned with LeapAlign using HPSv3 as the reward model.}
\label{fig:hpsv3_qualitative}
\end{figure*}

\begin{figure*}[t]
\centering
\includegraphics[width=0.84\textwidth]{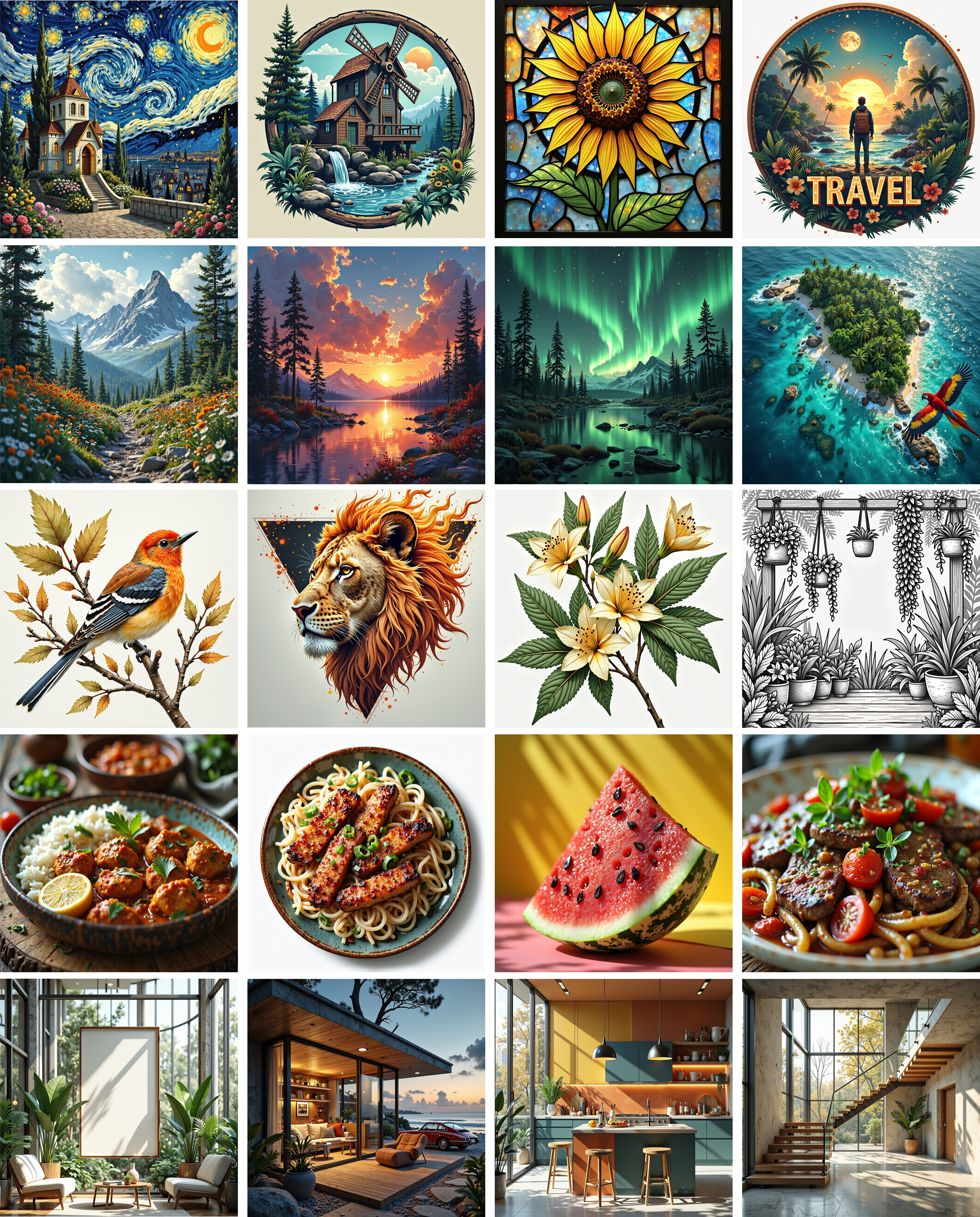}
\caption{Qualitative results of Flux fine-tuned with LeapAlign using HPSv3 as the reward model.}
\label{fig:hpsv3_qualitative2}
\end{figure*}

We present qualitative results of Flux fine-tuned with LeapAlign using HPSv3 as the reward model in Figures~\ref{fig:hpsv3_qualitative} and \ref{fig:hpsv3_qualitative2}. The fine-tuned model generates visually compelling and realistic images across diverse styles, themes, and scenarios, demonstrating that LeapAlign effectively aligns flow matching models with human preferences.

\end{document}